\documentclass[nohyperref]{article}

\usepackage{microtype}
\usepackage{graphicx}
\usepackage{booktabs} 

\usepackage{hyperref}

\usepackage[review]{icml2022}

\usepackage{amsmath}
\usepackage{amssymb}
\usepackage{mathtools}
\usepackage{amsthm}

\usepackage[capitalize,noabbrev]{cleveref}

\theoremstyle{plain}

\theoremstyle{definition}

\theoremstyle{remark}

\usepackage[textsize=tiny]{todonotes}

\usepackage{amsmath,amsfonts,bm}

\def\eqref#1{equation~\ref{#1}}
\def\Eqref#1{Equation~\ref{#1}}

\def\1{\bm{1}}

\DeclareMathAlphabet{\mathsfit}{\encodingdefault}{\sfdefault}{m}{sl}
\SetMathAlphabet{\mathsfit}{bold}{\encodingdefault}{\sfdefault}{bx}{n}

\usepackage{url}

\usepackage{amsfonts}
\usepackage{pifont}
\usepackage{nicefrac}
\usepackage{multirow}
\usepackage{subcaption}
\usepackage[font=small]{caption}
\usepackage{soul}
\usepackage{xcolor}
\usepackage{comment}
\usepackage{sidecap}
\usepackage{xspace}
\usepackage{enumitem}

\setlist[description]{font=\normalfont\itshape\space}

\usepackage[utf8]{inputenc}
\usepackage[T1]{fontenc}

\usepackage{tablefootnote}

\newcommand{\encoderonly}{\text{PrefixLM}\xspace}
\newcommand{\decoderonly}{\text{CausalLM}\xspace}
\newcommand{\encdec}{\text{EncDec}\xspace}
\newcommand{\lms}{\text{LM}s\xspace}
\newcommand{\lm}{\text{LM}\xspace}

\newcommand{\cmark}{\ding{51}}
\newcommand{\xmark}{\ding{55}}

\newcommand{\response}[1]{#1}

\icmltitlerunning{Examining Scaling and Transfer of Language Model Architectures for Machine Translation}

\begin{document}

\twocolumn[
\icmltitle{Examining Scaling and Transfer of Language Model Architectures \\ for Machine Translation}



\icmlsetsymbol{equal}{*}

\begin{icmlauthorlist}
\icmlauthor{Biao Zhang}{uoe,equal} 
\icmlauthor{Behrooz Ghorbani}{google}
\icmlauthor{Ankur Bapna}{google}
\icmlauthor{Yong Cheng}{google}
\icmlauthor{Xavier Garcia}{google}\\
\icmlauthor{Jonathan Shen}{google}
\icmlauthor{Orhan Firat}{google}
\end{icmlauthorlist}

\icmlaffiliation{google}{Google Research}
\icmlaffiliation{uoe}{School of Informatics, University of Edinburgh}

\icmlcorrespondingauthor{Biao Zhang}{b.zhang@ed.ac.uk}
\icmlcorrespondingauthor{Orhan Firat}{orhanf@google.com}

\icmlkeywords{Machine Learning, ICML}

\vskip 0.3in
]



\printAffiliationsAndNotice{$^*$Work done while interning at Google Research.}  



\begin{abstract}

Natural language understanding and generation models follow one of the two dominant architectural paradigms: \textit{language models} (\lms) that process concatenated sequences in a single stack of layers, and \textit{encoder-decoder models} (\encdec) that utilize separate layer stacks for input and output processing. In machine translation, \encdec has long been the favoured approach, but with few studies investigating the performance of \lms. 
In this work, we thoroughly examine the role of several architectural design choices on the performance of \lms on bilingual, (massively) multilingual and zero-shot translation tasks, under systematic variations of data conditions and model sizes. Our results show that: (i) Different \lms have different scaling properties, where architectural differences often have a significant impact on model performance at small scales, but the performance gap narrows as the number of parameters increases, (ii) Several design choices, including causal masking and language-modeling objectives for the source sequence, have detrimental effects on translation quality, and (iii) When paired with full-visible masking for source sequences, \lms could perform on par with \encdec on supervised bilingual and multilingual translation tasks, and improve greatly on zero-shot directions by facilitating the reduction of off-target translations.

\end{abstract}

\section{Introduction}

The popularity of large, general-purpose text generation models has skyrocketed in recent years due to their outstanding performance across a wide range of natural language processing (NLP) tasks~\citep{NEURIPS2020_1457c0d6, JMLR:v21:20-074,xue-etal-2021-mt5}. These generative models come in two flavors: encoder-decoder (\encdec) models~\citep{JMLR:v21:20-074} with two independent modules for encoding and decoding, and encoder-only ~\citep{devlin-etal-2019-bert} or decoder-only models \citep{{NEURIPS2020_1457c0d6}} that use a single module for both encoding and decoding functions (\lms). Often, these two types of architectures deliver comparable downstream performance under large-scale pretraining.

However, in neural machine translation (NMT), \encdec has been the dominant paradigm across all translation tasks (e.g. high/low-resource, multilingual and zero-shot translations)~\citep{barrault-etal-2020-findings,ansari-etal-2020-findings} with very few studies investigating the application of \lms~\citep{NEURIPS2018_4fb8a7a2,Wang2021LanguageMA}. Compared to \encdec, \lm offers a more compact architecture by sharing the parameters across encoding and decoding procedures. Considering these procedures are over distinct source/target languages for machine translation, sharing of the parameters across them implicitly affects the transfer dynamics: may result in improved representations by positive language transfer across languages \citep{DBLP:journals/corr/abs-1907-05019}, or may hurt the end-quality by amplifying capacity dilution problem \citep{lample2019crosslingual}. With concurrent streams of research in understanding the scaling properties of \lm~\citep{kaplan2020scaling} and \encdec~\citep{Ghorbani2021ScalingLF} paradigms, we see value in revisiting the NMT architecture inductive biases on a diverse set of translation tasks.  
 


\begin{figure*}[h]
    \centering
    \includegraphics[scale=0.60]{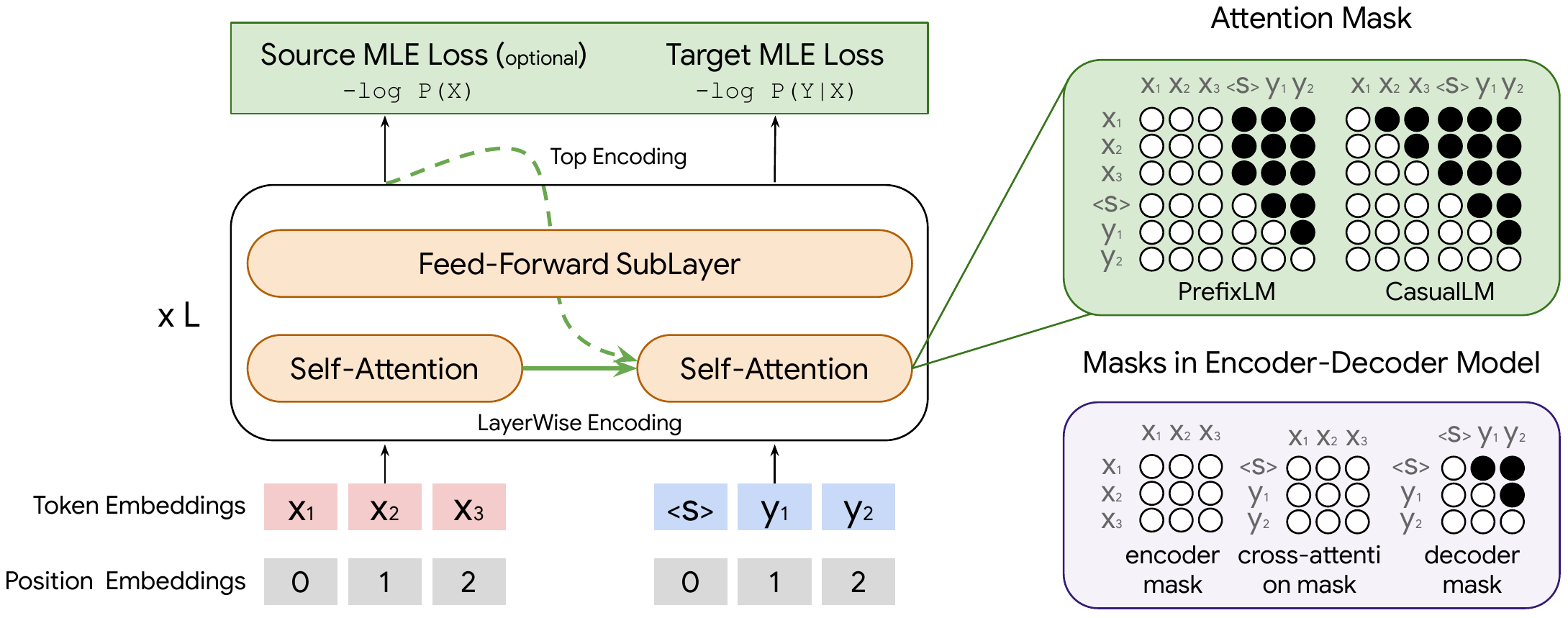}
    \caption{\label{fig:overview} Illustration for translation-oriented language models. $X$ and $Y$ denote source and target input, respectively. To enable translation, we adapt the \lm self-attention mask to either the \encoderonly mask or \decoderonly mask (top right), where filled black circles indicate disallowed attention. We also explore top-only encoding (Top Encoding) for \encoderonly which feeds the final-layer source encodings to generation similar to \encdec, rather than layer-wise coordinated encodings~\citep{NEURIPS2018_4fb8a7a2}. Masks of \encdec are shown in the bottom right for comparison.
    }
\end{figure*}

In this paper, we explore various configurations of \lm architectures for translation as illustrated in Figure \ref{fig:overview}. 
We compare them with the customary \encdec architecture along two axes, parameter scaling and cross-lingual transfer. We conduct a systematic study under a variety of data conditions, tasks (bilingual, multilingual and zero-shot) and examine recent architectural design choices associated with \lms, including causal masking (\decoderonly) vs. full-visible masking (\encoderonly) for source sequences,\footnote{Also known as unidirectional vs bidirectional language modelling, where in the unidirectional case a token representation takes into account only the preceding tokens and their representations, but the bidirectional case takes into account both preceding and following tokens in a sequence.} layer-wise coordination~\citep{NEURIPS2018_4fb8a7a2} vs. final-layer source encodings (TopOnly) for target sequence generation, increasing \lm depth vs. width, and also the effect of adding source language modeling loss for \decoderonly.

Our main findings are listed below:
\begin{itemize}
    \item \lms show different scaling properties compared to \encdec. The architectural differences become less important as models scale, measured by reduced quality gap against \encdec, regardless of the language similarities, training data conditions and evaluation settings.
    \item \encoderonly variants often outperform their \decoderonly counterparts; increasing \lm depth benefits the translation task more than increasing the width; and adding a source-side language modeling objective to \decoderonly does not yield significant translation quality gain.
    \item Cross-lingual transfer also benefits from model scaling, where \encdec almost always dominates the quality Pareto frontier on supervised directions while zero-shot translation favors \encoderonly and \lms. We also observed \encoderonly and \lms significantly reduce off-target translations.
\end{itemize}


\begin{table*}[t!]
\centering
\small
\caption{\label{tb:lm_variants} Comparison of different model variants studied in this paper. $X/Y$: source/target input. \textit{Layer-Wise}: layer-wise coordination ~\citep{NEURIPS2018_4fb8a7a2}; \textit{TopOnly}: use topmost-layer source encodings; \textit{Src-Src Mask}: the intra-source masking schema, either fully visible (Full) or causal (Causal); \textit{Parameter Sharing}: whether the parameters are shared during the processing of source and target sequences.}
\begin{tabular}{lcccccc}
\toprule
\multirow{2}{*}{Model} & \multicolumn{2}{c}{Objective} & \multicolumn{2}{c}{Structure} & \multirow{1}{*}{Src-Src} & \multirow{1}{*}{Parameter} \\
\cmidrule(lr){2-3} \cmidrule(lr){4-5}
& $-\log P(X)$ & $-\log P(Y|X)$ & Layer-Wise & TopOnly & Mask & Sharing \\
\midrule
\encdec & & \cmark & & \cmark & Full & \xmark \\
\midrule
\encoderonly & & \cmark & \cmark & & Full & \cmark \\
\quad + TopOnly & & \cmark & & \cmark & Full & \cmark \\
\midrule
\decoderonly & \cmark & \cmark & \cmark & & Causal & \cmark \\
\quad + TgtOnly & & \cmark & \cmark & & Causal & \cmark \\
\bottomrule
\end{tabular}
\end{table*}
\section{Related Work}

Using language models in the task of translation has a long history, particularly in the era of statistical machine translation (SMT) where \lm was used as a separate yet crucial component ensuring the fluency of generation~\citep{Stolcke2002SRILMA,heafield-2011-kenlm,10.5555/1734086}. With neural networks, NMT unified those isolated SMT components including \lm under the encoder-decoder formulation~\citep{kalchbrenner-blunsom-2013-recurrent,cho2014learning,DBLP:journals/corr/SutskeverVL14,bahdanau+al-2014-nmt}, which makes use of separate modules to process input and output.
Further studies exploring architectural modifications by using \lm alone as a translation model, nevertheless, got much less attention. \citet{NEURIPS2018_4fb8a7a2} proposed \textit{layer-wise coordination}  between encoder and decoder with tied weights, where each decoder layer attends to its corresponding encoder layer at the same depth as opposed to the conventional method of attending the top-most encoder representations. Later, \citet{Fonollosa2019JointSS} extended it with locality constraint. \citet{NEURIPS2019_c20bb2d9} explored \lms for sequence generation under large-scale pretraining. Despite reporting promising results, these prior studies either focus only on bilingual tasks or do not consider the scaling properties of the models, leaving the picture incomplete: how the findings will change as we scale the models and how the languages benefit from/interfere each other as the architectural priors (inductive biases) change.


Neural models follow some scaling laws. \citet{kaplan2020scaling} reported the test cross-entropy loss of \lms can be formulated as a power-law scaling function of either model size (excluding embedding parameters) or dataset size. Later on, researchers examined and confirmed such findings across different domains, including vision modeling~\citep{Zhai2021ScalingVT}, knowledge transfer from pretraining~\citep{Hernandez2021ScalingLF}, autoregressive generative modeling~\citep{Henighan2020ScalingLF}, and neural machine translation~\citep{gordon2021data,Ghorbani2021ScalingLF}, to name a few. We find it essential to study the scaling behavior of new architectures and approaches given the recent evidence on the emergent properties of the models at scale \citep{NEURIPS2020_1457c0d6}.

Another critical component in machine translation is the number of languages being considered with the models, which is the very focus of multilingual NMT \citep{firat2016multi}. Cross-lingual transfer in multilingual NMT often results from parameter sharing across languages, which benefits low-resource languages and also enables zero-shot translation~\citep{johnson-etal-2017-googles}, although the quality on zero-shot directions is largely hindered by the off-target translation problem~\citep{DBLP:journals/corr/abs-1903-07091,zhang-etal-2020-improving}. The structure of \lms further encourages parameter sharing, offering a chance to improve the transfer while magnifying the problem of interference (negative-transfer)~\citep{wang-etal-2020-negative,zhang2021share}. Very recently, \citet{Wang2021LanguageMA} analyzed the cross-lingual transfer behavior of \decoderonly, and reported encouraging zero-shot performance. However, we did not observe the same results likely because of data sampling, model architecture and optimization differences which zero-shot transfer is sensitive to.

\section{Language Model Architectures for MT}

In this section, we first briefly review \encdec and then present \lm architectures for translation based on Transformer~\citep{NIPS2017_7181}. Table \ref{tb:lm_variants} shows different models. 
Given a source sequence $\mathbf{X}$ of length $|X|$ and its target translation $\mathbf{Y}$ of length $|Y|$, \encdec performs translation via the following structure:
\begin{equation}\label{eq:encdec_structure}
    \begin{split}
    \mathbf{X}^l = & \text{FFN}\circ\text{SAtt}\left(\mathbf{X}^{l-1}\right), \\
    \mathbf{Y}^l = & \text{FFN}\circ\text{CAtt}\circ\text{SAtt}\left(\mathbf{Y}^{l-1}, \mathbf{X}^L\right),
    \end{split}
\end{equation}
where $l$ denotes the layer index and $\circ$ indicates consecutive sublayers. $\mathbf{X}^{l} \in \mathbb{R}^{|X|\times d}$ and $\mathbf{Y}^{l} \in \mathbb{R}^{|Y|\times d}$ are the layer representations of the source and target sequence respectively, with a model dimension of $d$. The first input layer ($\mathbf{X}^0, \mathbf{Y}^0$) is the summation of token embeddings and their positional encodings. {We drop all the layer normalization and residual connections in our formulations for brevity.}

The encoder is a stack of $L$ layers, each of which includes a multi-head self-attention sublayer (SAtt) followed by a feed-forward sublayer (FFN). SAtt in the encoder is bidirectional with \textit{full-visible masking} that has full visibility to all source tokens, preceding and following. Its final-layer representations $\mathbf{X}^L$ are fed to the decoder, which shares a similar structure to the encoder but with an additional (multi-head) cross-attention sublayer (CAtt). Unlike encoder, SAtt in the decoder is unidirectional with \textit{causal masking}, where attention to following tokens is disabled (masked). CAtt can always access all source inputs, though. Note we set the encoder and decoder depth equally to $L$, and use $d^{\text{ff}}$ to denote the intermediate dimension of FFN. \encdec is often optimized with the target translation objective based on $\mathbf{Y}^L$:
\begin{equation}\label{eq:encdec_objective}
    \mathcal{L}^{\encdec}(X, Y) = \mathcal{L}^{\text{TGT}} = -\log P(Y|X, \mathbf{Y}^L).
\end{equation}

Instead of separately modeling source and target sequences, \lm handles both with a single module:
\begin{equation}\label{eq:lm_structure}
    \left[\mathbf{X}^l, \mathbf{Y}^l\right] = \text{FFN}\circ\text{SAtt}\left(\left[\mathbf{X}^{l-1}, \mathbf{Y}^{l-1}\right], \mathbf{M}\right),
\end{equation}
where $\mathbf{M} \in \{0, 1\}^{(|X|+|Y|)\times(|X|+|Y|)}$ is the attention mask that controls the information flow within the concatenated sequences ($[\cdot, \cdot]$).\footnote{Note that, in our implementation we still use separate source and target positions as shown in Figure \ref{fig:overview}.} Two \lm variants explored by changing the structure of mask $\mathbf{M}$, \textit{\encoderonly} and \textit{\decoderonly}.

\noindent{\bf \encoderonly} merges different modules of \encdec, trained with $\mathcal{L}^{\text{TGT}}$. Its attention mask
\begin{equation}\label{eq:enconly_mask}
    \mathbf{M}^{\encoderonly}(i, j) = 1, \text{ if } i \geq j \text{ or } j \leq |X| \text{; otherwise } 0,
\end{equation}
combines the encoder/decoder self-attention mask and the cross-attention mask of \encdec. $1 \leq i, j $ $\leq |X|+|Y|$, and masks of value 0 mark the attention as unavailable.

\noindent{\bf\decoderonly}, by contrast, is a strict \lm that applies causal masking to both sequences:
\begin{equation}\label{eq:deconly_mask}
    \mathbf{M}^{\decoderonly}(i, j) = 1, \text{ if } i \geq j \text{; otherwise } 0.
\end{equation}

\begin{table*}[t!]
\centering
\small
\caption{\label{tb:dataset} Statistics of different datasets. \textit{M/B}: million/billion; \textit{SO/TO}: source-original/target-original test sets; \textit{Web}: in-house web-crawled datasets; \textit{BIL/MUL}: the data is used for bilingual/multilingual experiments.}
\begin{tabular}{lcccccc}
\toprule
\multirow{2}{*}{Dataset} & \multicolumn{3}{c}{\#Samples (Sources)} & \multicolumn{2}{c}{Experiments} \\
\cmidrule(lr){2-4} \cmidrule(lr){5-6}
& Train & Dev & Test & BIL & MUL \\
\midrule
WMT14 En-De & 4.5M & 3000 (WMT13) & 3003 (WMT14) & & \cmark \\
WMT14 En-Fr & 41M & 3000 (WMT13) & 3003 (WMT14) & \cmark & \cmark \\
WMT19 En-Zh & 26M & 3981 (WMT18) & \begin{tabular}{@{}c@{}} 1997 (WMT19, SO) \\ 2000 (WMT19 TO) \end{tabular} & \cmark & \cmark \\
Web En-De & 2B & 7927 (Web) & \begin{tabular}{@{}c@{}} 4927/1997 (Web/WMT19, SO) \\  6000/2000 (Web/WMT19, TO)\end{tabular} & \cmark & \\
\bottomrule
\end{tabular}
\end{table*}

Apart from $\mathcal{L}^{\text{TGT}}$, \decoderonly also includes the source-side language modeling loss for training:
\begin{align}
        \mathcal{L}^{\decoderonly}(X, Y) & = \mathcal{L}^{\text{SRC}} + \mathcal{L}^{\text{TGT}} \label{eq:deconly_objective} \\
            & = - \log P(X | \mathbf{X}^L) - \log P(Y |X, \mathbf{Y}^L). \nonumber
\end{align}

To improve our understanding of \lms for translation, we further incorporate two extensions:
\begin{description}
    \item[\encoderonly + TopOnly] The model defined in \Eqref{eq:lm_structure} performs attention over the source and target sequence within the same layer. In contrast, \encdec always uses the topmost-layer source encodings for translation. 
    We mimic this with the \textit{TopOnly} extension by feeding top-layer encodings, i.e. $\mathbf{X}^L$ instead of $\mathbf{X}^{l-1}$, to each attention sublayer. It operates the same as \encdec but with the parameters of encoder and decoder tied.
    
    \item[\decoderonly + TgtOnly] The inclusion of the source-side objective enriches \decoderonly's learning signal and encourages the model to absorb source language characteristics. However, it requires and occupies part of modeling capacity, which might negatively affect translation. To offset this impact, we add the \textit{TgtOnly} extension that optimizes \decoderonly with the target translation objective $\mathcal{L}^{TGT}_C$ alone, which also aligns better with \encdec and \encoderonly.
\end{description}

\section{Setup}

\paragraph{ Model Setting} We use Transformer for experiments. By default, we adopt the base setting, with $d=512$, $d^{\text{ff}}=2048$ and 8 attention heads. We also work with the Transformer big setting where each hyper-parameter above is doubled. Training and inference details are in Appendix \ref{app:train_inference}.

\paragraph{Datasets and Evaluation} We use WMT14 English-French (En-Fr), WMT14 English-German (En-De), WMT19 English-Chinese (En-Zh) and an in-house web-crawled (Web) En-De dataset for experiments, whose statistics are summarized in Table \ref{tb:dataset}. We also report results on OPUS-100~\citep{zhang-etal-2020-improving}, a massively multilingual corpus containing 100 languages. All datasets are pre-processed with byte pair encoding~\citep[BPE]{sennrich-etal-2016-neural} implemented by SentencePiece~\citep{kudo-richardson-2018-sentencepiece}. We set the BPE vocabulary size to 32K by default. We report test log-perplexity score (PPL) for scaling study particularly and also show SacreBLEU~\citep{post-2018-call}\footnote{Signature: \textit{BLEU+c.mixed+lang*+\#r.1+s.exp+t.13a+v.1.5.1}}.



\section{Experiments for Model Scaling}

\citet{kaplan2020scaling} reported that the model performance can be described with a power-law, with respect to its parameters, as below:
\begin{equation}\label{eq:scaling_law}
    \mathcal{L}(N) = \alpha \left(\frac{N_0}{N}\right)^p + \mathcal{L}_{\infty},
\end{equation}
where $\mathcal{L}(N)$ fits test PPL, and $N$ denotes the number of parameters. $N_0$ is a constant used for numerical stability which is obtained from 1-layer \encdec model. $\alpha, p, \mathcal{L}_{\infty}$ are \textit{fitted} parameters, and we mainly analyze the estimated scaling exponent $p$ and the irreducible loss $\mathcal{L}_{\infty}$. 

\begin{figure*}[t!]
    \centering
    \subcaptionbox{\label{fig:wmt_enfr_base_scaling} Log-Perplexity for En$\rightarrow$Fr}{
        \begin{minipage}[t]{0.24\textwidth}
        \centering
        \includegraphics[scale=0.325]{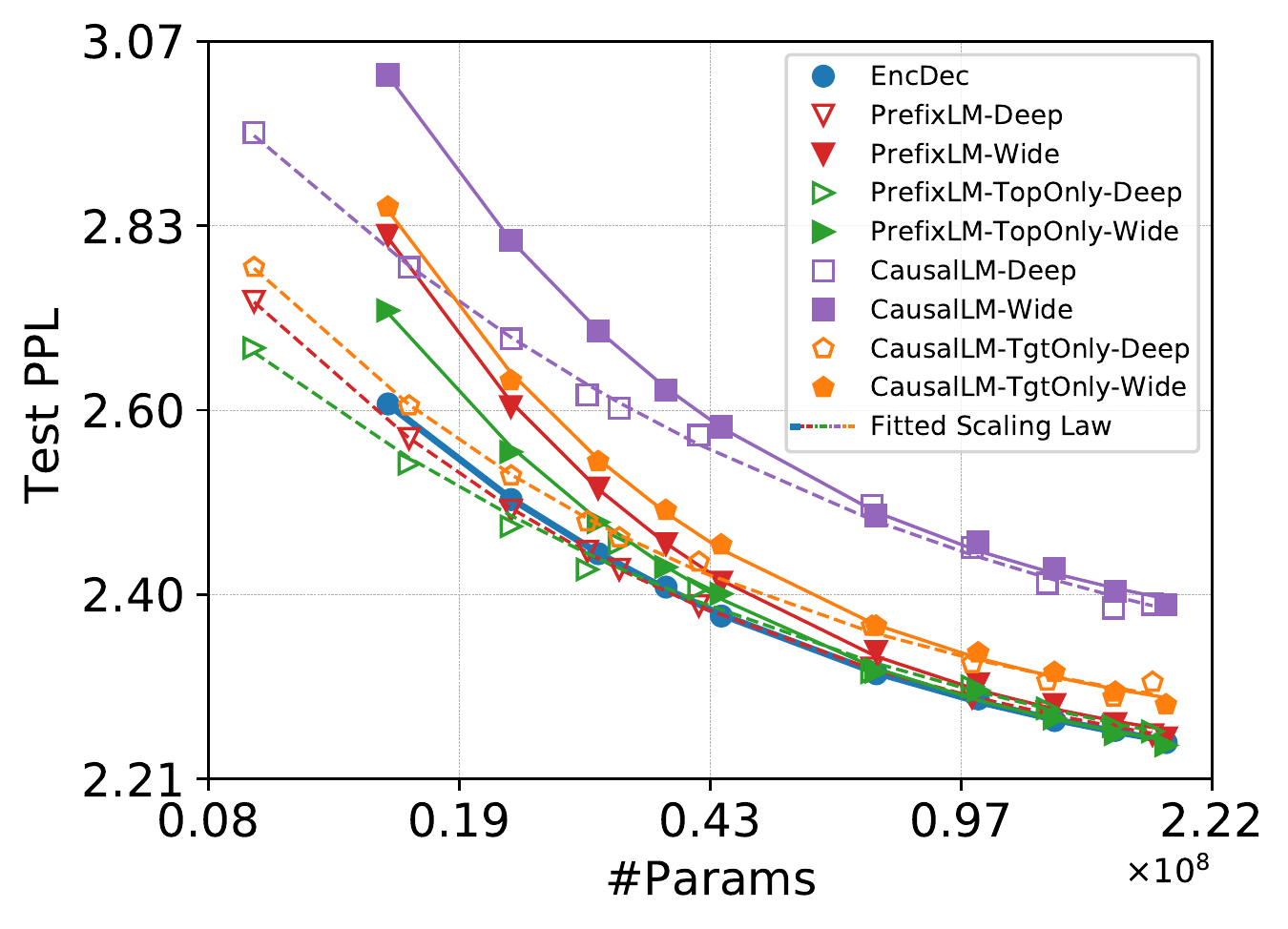}
        \end{minipage}%
    }
    \subcaptionbox{\label{fig:wmt_enzh_base_scaling} Log-Perplexity for En$\rightarrow$Zh}{
        \begin{minipage}[t]{0.24\textwidth}
        \centering
        \includegraphics[scale=0.325]{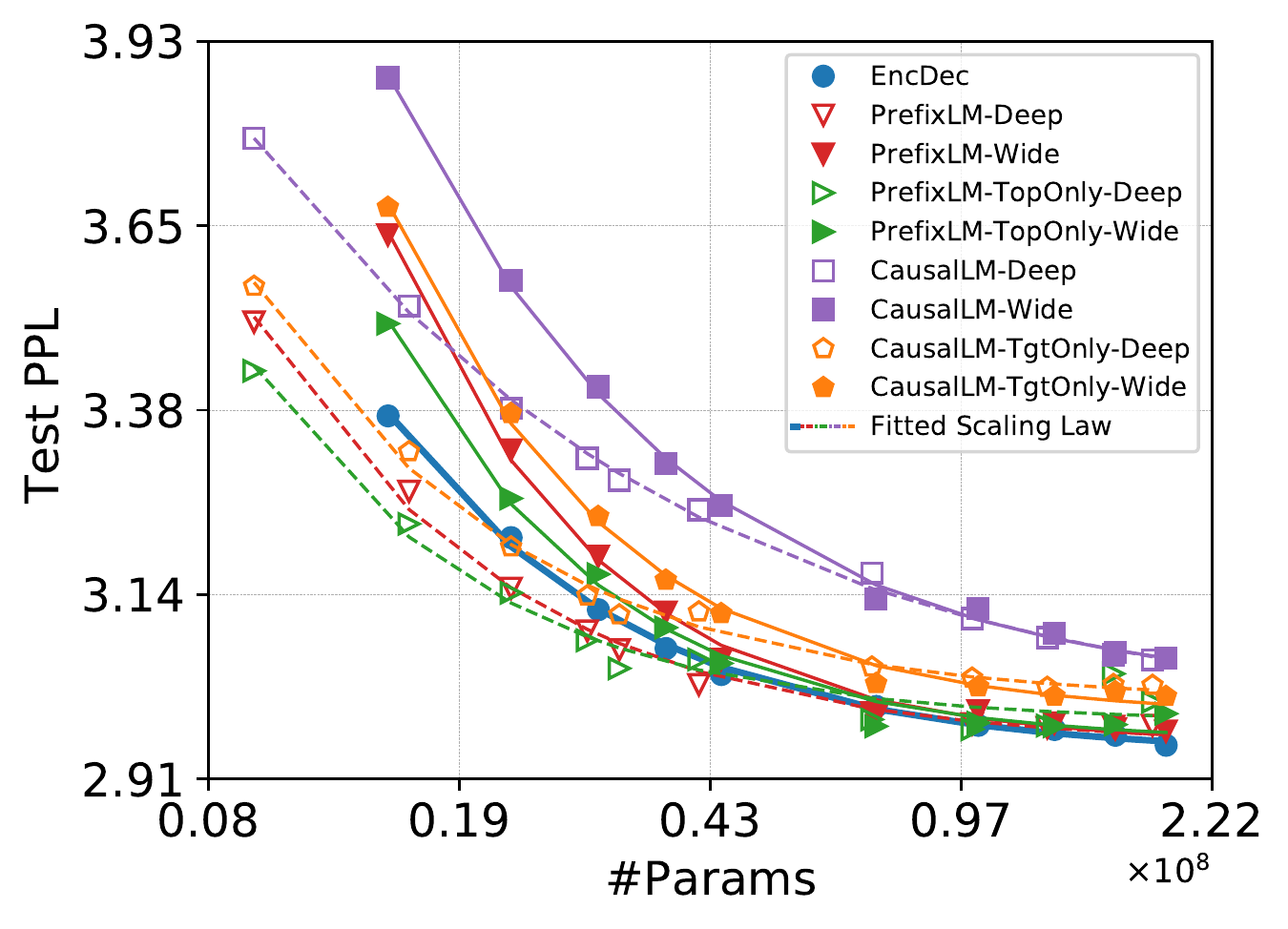}
        \end{minipage}%
    }
    \subcaptionbox{\label{fig:wmt_enfr_base_scaling_bleu} BLEU for En$\rightarrow$Fr}{
        \begin{minipage}[t]{0.24\textwidth}
        \centering
        \includegraphics[scale=0.325]{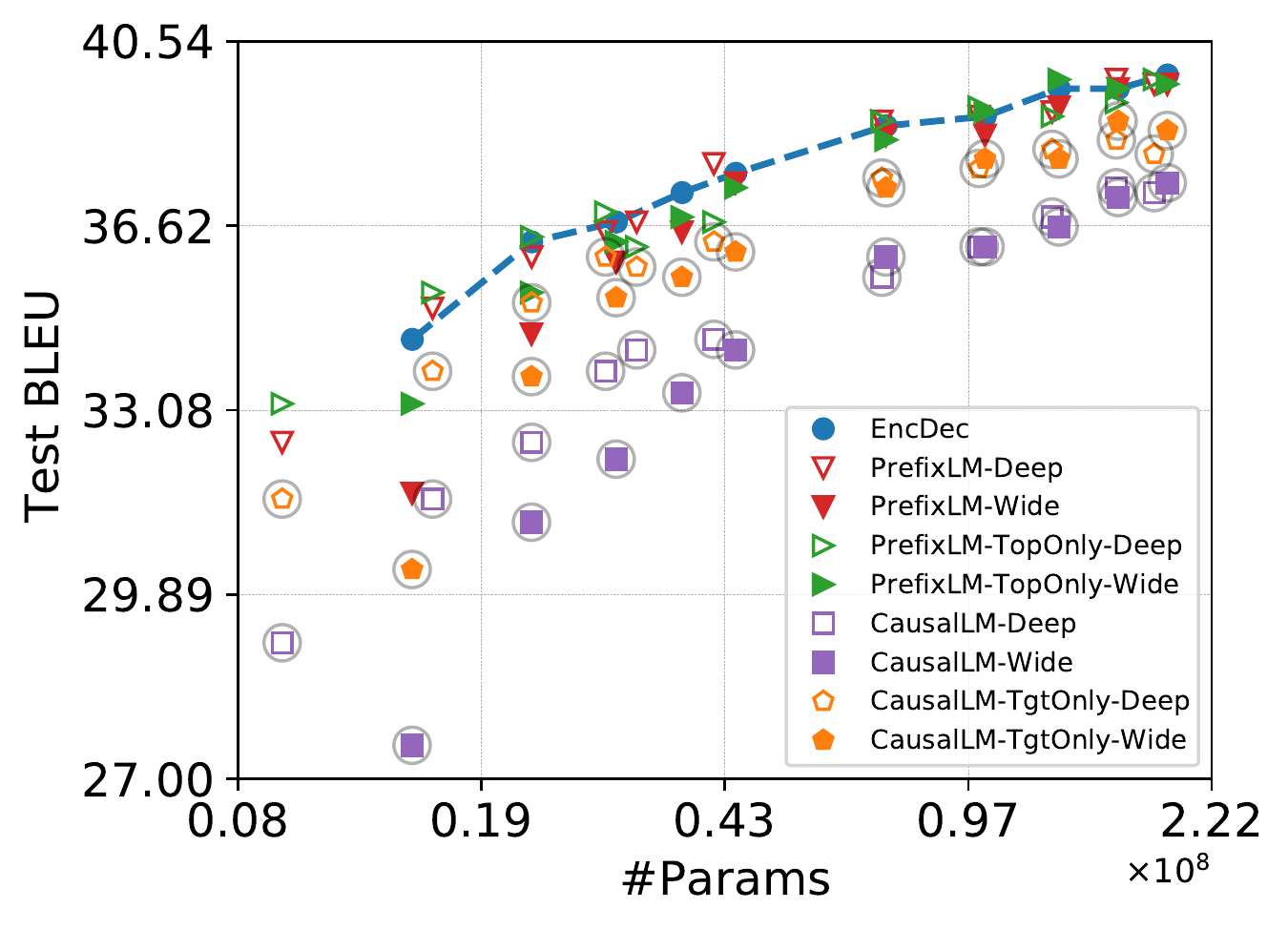}
        \end{minipage}%
    }
    \subcaptionbox{\label{fig:wmt_enzh_base_scaling_bleu} BLEU for En$\rightarrow$Zh}{
        \begin{minipage}[t]{0.24\textwidth}
        \centering
        \includegraphics[scale=0.325]{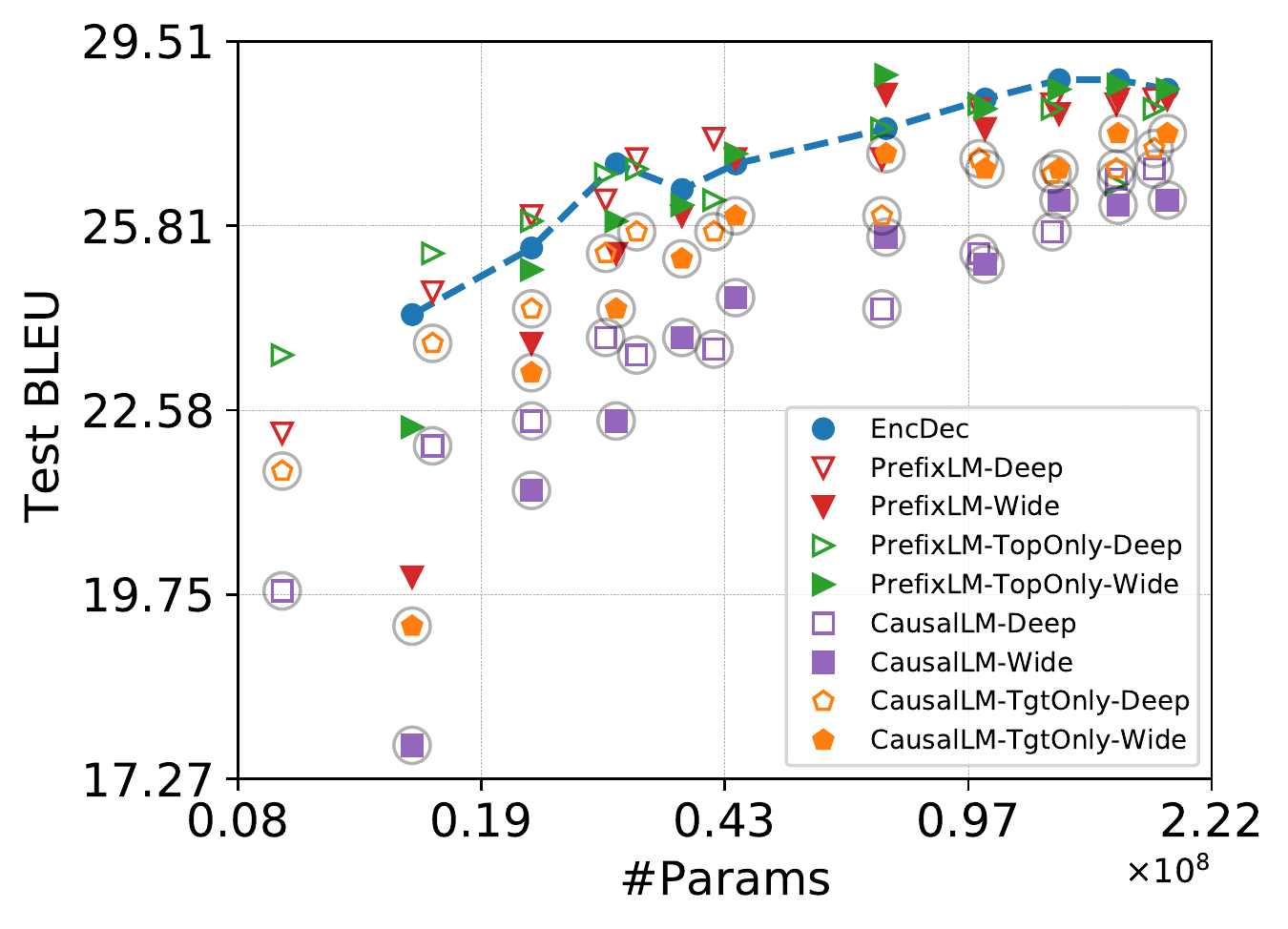}
        \end{minipage}%
    }
    \caption{\label{fig:base_scaling} Fitted scaling curves (\ref{fig:wmt_enfr_base_scaling},\ref{fig:wmt_enzh_base_scaling}) and BLEU scores (\ref{fig:wmt_enfr_base_scaling_bleu},\ref{fig:wmt_enzh_base_scaling_bleu}) for different models on WMT14 En-Fr (left) and WMT19 En-Zh (right) tasks. \textbf{Top}: dashed and solid fitted curves are for \textit{\lm + Deep} and \textit{\lm + Wide}, respectively. We represent the \encdec scaling with bold solid curve. \textbf{Bottom}: dashed curve denotes the BLEU scores of \encdec as a function of model parameters for reference. Markers in circles are for \decoderonly variants. Models are trained in Transformer base setting. Best seen in color.}
\end{figure*}

\begin{figure*}[t]
    \centering
    \includegraphics[scale=0.38]{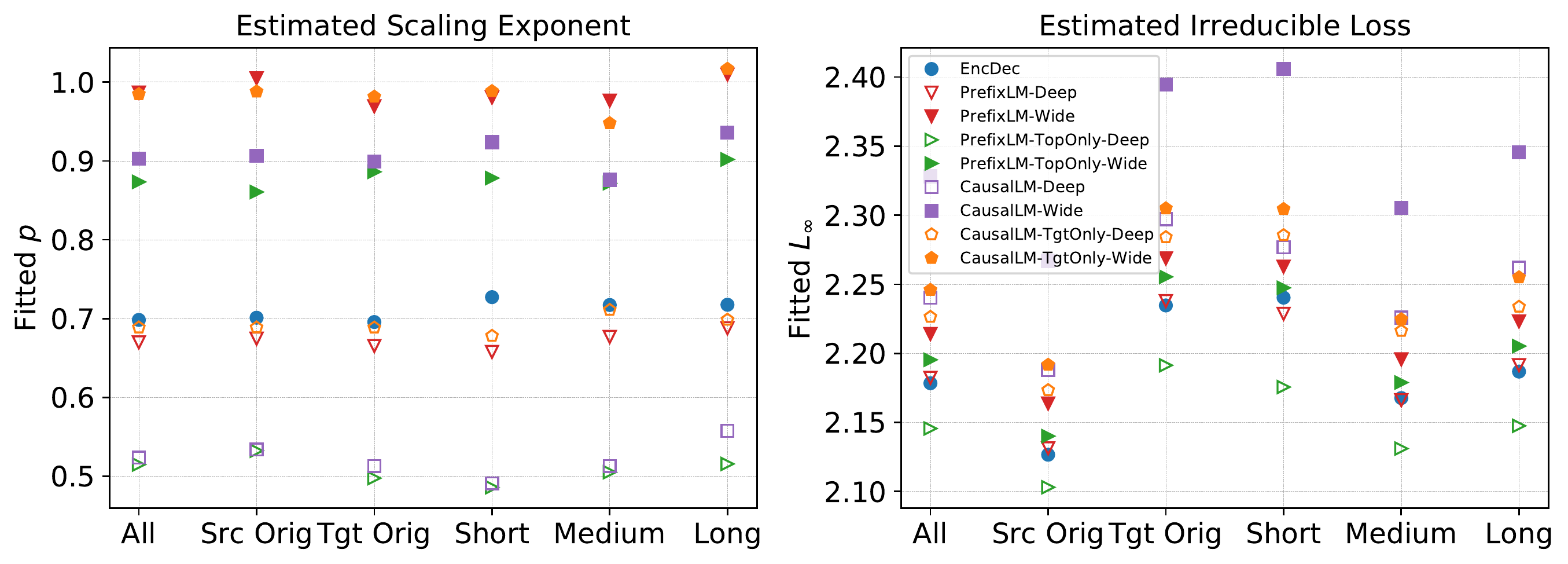}
    \caption{\label{fig:base_scaling_exp_irreduc} Fitted scaling exponent ($p$, left) and irreducible loss ($\mathcal{L}_{\infty}$, right) over different evaluation settings on WMT14 En-Fr (En$\rightarrow$Fr). \textit{All}: the whole test set; \textit{Src Orig}, \textit{Tgt Orig}: source-original and target-original test set, respectively; \textit{Short, Medium, Long}: shortest, medium and longest $\sim$376 samples from the test set, respectively.}
\end{figure*}

The way of increasing model parameters varies for the same model and also across different models. We perform scaling firstly for \encdec by changing its depth $L$ (from 1 to 26 layers, equally for its encoder and decoder) while keeping the other hyper-parameters intact following~\citet{Ghorbani2021ScalingLF}. We then align the scaling settings of \lm with its \encdec counterpart in term of model parameters through increasing either its depth or width:
\begin{description}
    \item [\lm + Deep] adds parameters by stacking more Transformer layers, which was also used in previous studies~\citep{NEURIPS2018_4fb8a7a2,Wang2021LanguageMA}. 
    \item [\lm + Wide] instead, grows the model width. We choose to enlarge the feed-forward dimension from $d^{\text{ff}}$ to $3d^{\text{ff}}$. Note other strategies for width scaling are possible and many, but exploring them is resource-consuming and beyond the scope of our paper.
\end{description}

We distinguish data-limited regime from model size-limited regime for model scaling~\citep{Bahri2021ExplainingNS}, where the former has relatively fewer training samples than model parameters thus likely suffers from overfitting (e.g. with WMT14 En-Fr and WMT19 En-Zh), while the latter has enough samples for model fitting (e.g. with Web En-De).

\subsection{Scaling in Data-Limited Regime}

\paragraph{Architectural difference matters most when the model is at a small scale.} Figure \ref{fig:base_scaling} summarizes the scaling results on WMT14 En-Fr and WMT19 En-Zh. When there are fewer parameters, the model with inductive biases favoring translation achieves better quality. Such inductive bias includes 1) allowing the full visibility to the source input as in \encoderonly\footnote{By default, we use \encoderonly (\decoderonly) to refer to all \encoderonly variants (\decoderonly variants). We adopt the \textit{italic} form to denote a specific variant.} rather than causal masking; 2) using topmost-layer source encodings for translation (\textit{TopOnly}) rather than layer-wise coordinated encodings; 3) deeper \lms (\textit{Deep}) rather than wider models; and 4) training \lms without source-side language modeling loss (\textit{TgtOnly}). The fact that \textit{\lm + Deep} outperforms \textit{\lm + Wide} demonstrates that not only the number of parameters matters, but also the way parameters are added. This aligns with the previous findings: deeper models apply more non-linear operations and induce more abstract representations, which often improves translation quality~\citep{wang-etal-2019-learning-deep}. This also applies to TopOnly. Most of these findings are consistent across different languages and evaluation metrics.

We argue that factors making the \textit{TopOnly} variant favorable to translation tasks could be plenty. Based on the literature~\cite{tenney-etal-2019-bert}, representations in Transformer often evolve from the bottom up, where lower-layer encodings align better with syntactic-related information while the higher-layer representations correlate more with semantic-related information \citep{kudugunta2019investigating}. Given that the task of language translation is requires source-side semantic knowledge to provide clues for accurate source-target alignment, we speculate that the top-most source encodings could be preferred while generating the target sequence. Which has plausibility to explain the narrowed performance gap between Deep and TopOnly-Deep, since deeper layers could offer more abstract and semantic-intensive representations to the decoder to ensure the translation accuracy. 

\paragraph{Different models show different scaling properties, but the gap narrows at scale.} The impact of added parameters on translation quality differs across different models. The \lms that perform poorly at small scales often gain more from the increased capacity via adding parameters. For instance, the difference between \textit{\lm + Deep} and \textit{\lm + Wide} almost disappears at the end, resonating with the optimal depth-vs.-width theory~\citep{NEURIPS2020_ff4dfdf5}.
We observe that \encoderonly and \encdec converge to a similar quality bands followed by \textit{\decoderonly + TgtOnly} while \textit{\decoderonly} still retains a clear gap against the others. This performance gap is smaller in WMT19 En-Zh, mainly because of model overfitting. BLEU scores in Figure \ref{fig:wmt_enfr_base_scaling_bleu} and \ref{fig:wmt_enzh_base_scaling_bleu} also show similar trends, although the relationship between BLEU and PPL is non-trivial~\citep{Ghorbani2021ScalingLF}. These tell us that the success of architectural modifications on small-scale models may not transfer to large-scale settings, and that comparing different models under one model configuration in terms of the scale risks the results to be inconclusive.
Note we also observe reduced gap when considering the number of layers (see Figure \ref{fig:base_scaling_layer} in the Appendix).

\begin{figure}[t!]
    \centering
    \includegraphics[scale=0.40]{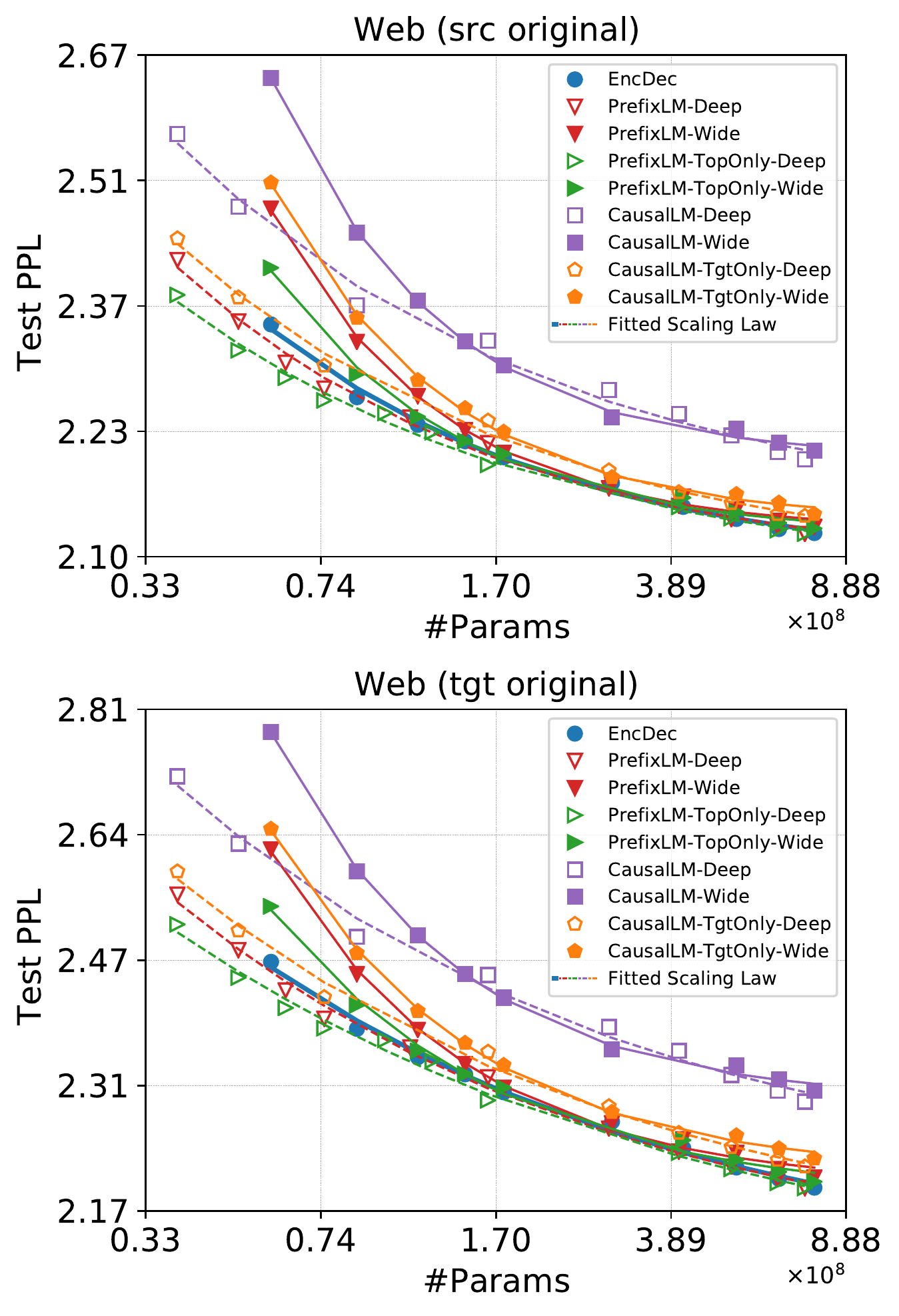}
    \caption{\label{fig:base_scaling_prod_part} Fitted scaling curves for different models on Web En-De (En$\rightarrow$De). \textit{src/tgt}: source/target; \textit{Web}: in-domain evaluation set. Models are trained in the Transformer big setting.}
\end{figure}

\begin{figure*}[t]
    \centering
    \includegraphics[width=0.92\textwidth]{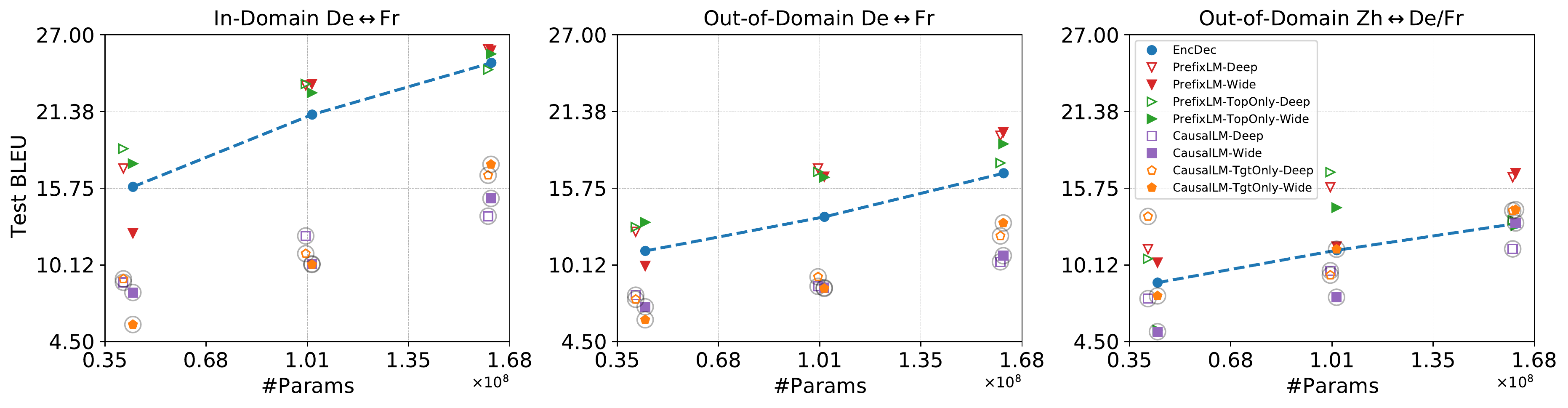}    
    \includegraphics[width=0.92\textwidth]{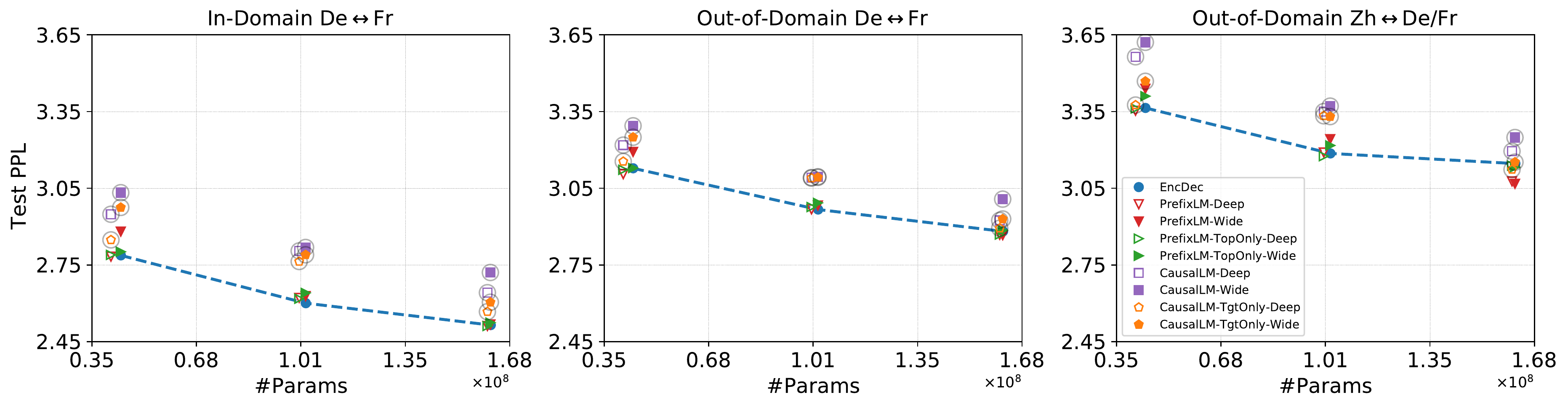}    
    \includegraphics[width=0.92\textwidth]{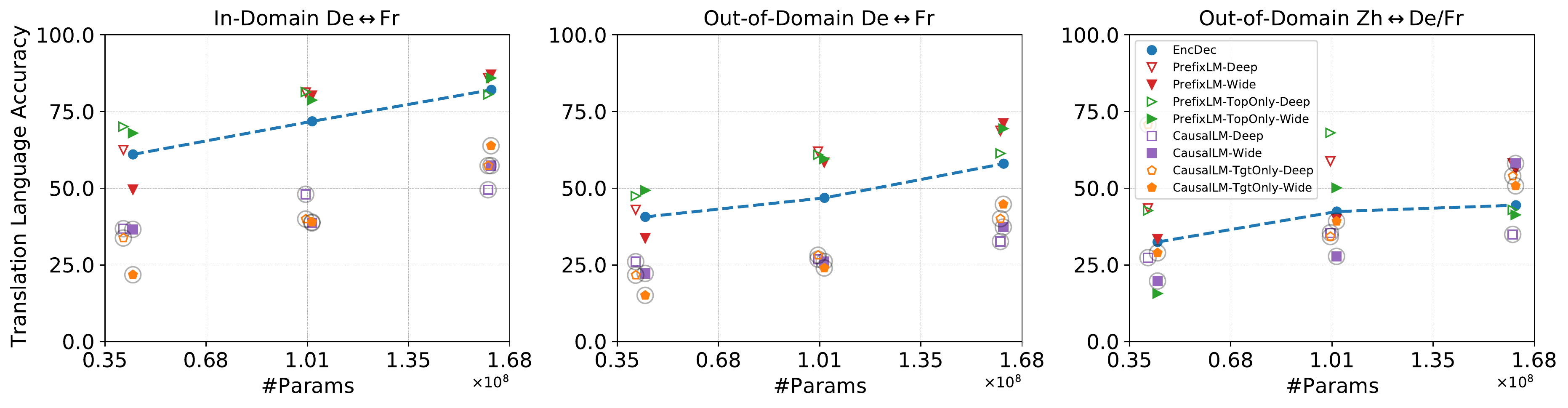}    
    \caption{\label{fig:transfer_zero_shot} Zero-shot transfer results of different models for multilingual many-to-many modeling on four languages (En-De-Fr-Zh) under different model sizes. \textbf{Top}: average BLEU scores; \textbf{Middle}: average PPL scores; \textbf{Bottom}: average translation language accuracy scores. \textit{In-domain}: WMT test set; \textit{Out-of-domain}: in-house sport-domain test sets.}
\end{figure*}
\begin{figure*}[t]
    \centering
    \includegraphics[scale=0.40]{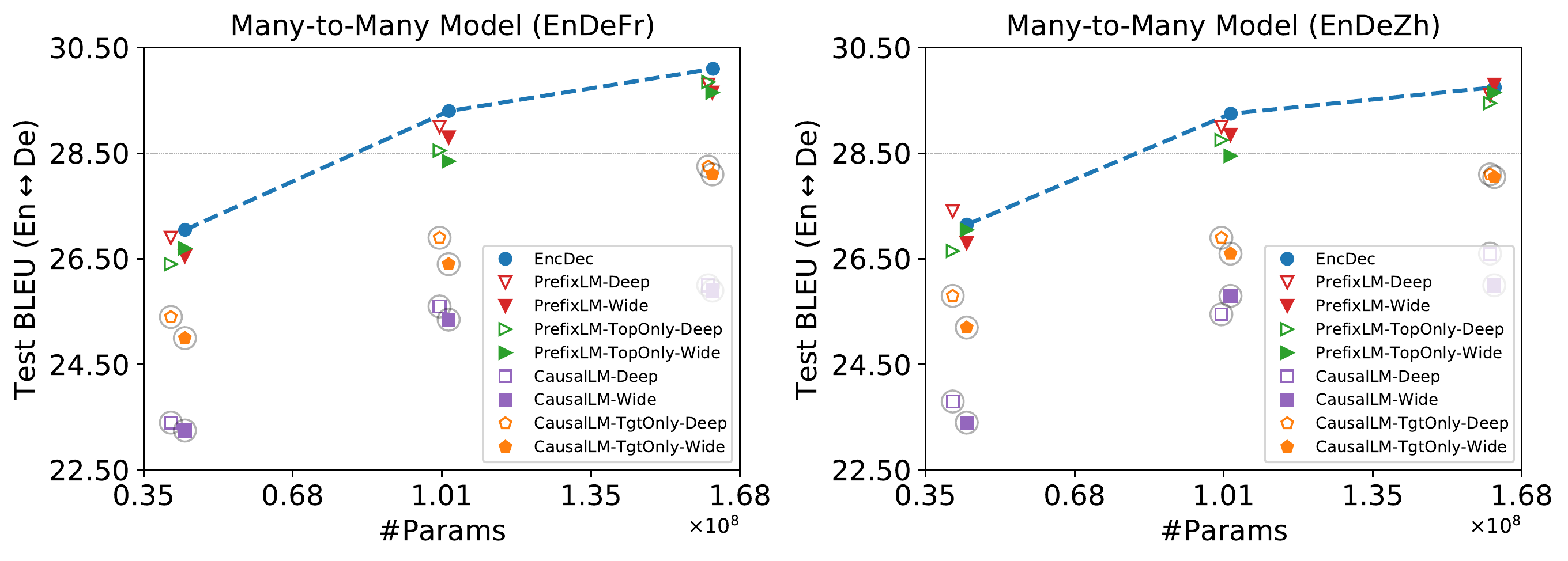}
    \caption{\label{fig:transfer_high_to_low} Cross-lingual transfer results (average BLEU scores) for different models from high-resource languages to the low-resource one (En-De) under different model sizes on WMT datasets. Average is performed over En$\rightarrow$De and De$\rightarrow$En evaluation. \textbf{Left}: multilingual En-De-Fr system; \textbf{Right}: multilingual En-De-Zh system. Both systems are many-to-many models. Models are trained in the base setting.}
\end{figure*}

\paragraph{Sequence lengths and the \textit{originality} of the test set affect does not affect scaling properties} We further test how the scaling changes across different evaluation settings, and show the results on WMT14 En-Fr in Figure \ref{fig:base_scaling_exp_irreduc}. The scaling exponent changes marginally over different settings (often less than 0.05), suggesting that the scaling curves are quite similar in these settings (see Figure \ref{fig:base_scaling_long}, \ref{fig:base_scaling_srctgt}, \ref{fig:base_scaling_614_bleu} in Appendix), although sentences of different originalities differ largely in style and naturalness~\citep{graham-etal-2020-statistical,freitag-etal-2020-bleu}. The estimated irreducible loss shows that target-original parallel sentences are harder to model than the source-original ones, and that translating medium-length sequences is much easier. The loss ranking of different models changes little over these settings, supporting \encoderonly and \encdec generally more than \decoderonly.

\paragraph{Computational efficiency favors \encdec over all \lms variants} Our calculation of FLOPs for different architectures show that \encdec models demand generally less computation compared to \lm, but the gap narrows at scale. Note \lm does not save any computations because of the quadratic attention over the concatenated source and target sequences. By contrast, to perform similarly to \encdec, \lm often needs to be made wider or deeper, which further deteriorates the computational efficiency both during training and inference time. Besides, \encdec allows arbitrary decoders, e.g. shallow decoders for faster inference, which is non-feasible for \lms. We also observed adding the source-side loss hurts \decoderonly's efficiency. We share the details of computational efficiency in Appendix, Figure \ref{fig:base_scaling_flops}.

\subsection{Scaling in Model Size-Limited Regime}

Figure \ref{fig:base_scaling_prod_part} shows the in-domain scaling performance on Web En-De. Overall, we observe similar scaling patterns as reported above, and such pattern transfers to out-of-domain evaluation, FLOPs and BLEU scores. More results are available in the Appendix (Figure \ref{fig:base_scaling_prod}, \ref{fig:base_scaling_prod_flops} and \ref{fig:base_scaling_prod_bleu}).

\section{Experiments for Cross-Lingual Transfer}

Based on previous studies~\citep{wang-etal-2020-negative,zhang2021share}, sharing capacity across languages could encourage knowledge transfer but might also gain the risk of negative interference. In this section, we further compare different models but on multilingual many-to-many translation. To enable multilingual NMT, we append a target language tag to each source sentence following~\citet{johnson-etal-2017-googles}. We perform over-sampling to balance the training data with a temperature of $T=5$~\citep{DBLP:journals/corr/abs-1907-05019}.

\paragraph{\encoderonly benefits zero-shot transfer.} We start with multilingual translation for WMT En-De/Fr/Zh, and regard En-De as a relatively low-resource language pair. We test how \lms perform on zero-shot translation. We use the newstest2019 De-Fr test set as the in-domain zero-shot eval set, and an internal sports-domain N-way test set for De-Fr-Zh (2000 samples) as the out-of-domain eval set. Figure \ref{fig:transfer_zero_shot} shows the results. Scaling improves knowledge transfer for almost all models, while \encoderonly performs surprisingly well on zero-shot directions. In most settings, \encoderonly surpasses \encdec significantly with respect to BLEU, and such superiority is more obvious on out-of-domain evaluation and for distant language pairs. 

Nevertheless, we find that \encoderonly usually underperforms \encdec in terms of PPL. In other words, \encdec still possesses the best fitting ability on zero-shot language pairs. Results on translation language accuracy explains this mismatch: compared to \encdec, \encoderonly drastically reduces off-target translation -- a bottleneck of zero-shot translation~\citep{zhang-etal-2020-improving}. This also suggests that \encdec suffers from more serious searching errors during inference~\citep{stahlberg-byrne-2019-nmt}, which the inductive biases of \encoderonly help. 

In addition, we observe no benefits from \decoderonly on zero-shot translation, with or without the source-side language modeling objective. This finding disagrees with that of~\citet{Wang2021LanguageMA}, which we ascribe to various differences in model, data and optimization. Note that \citet{Wang2021LanguageMA} adopted more aggressive data oversampling, didn't consider distant languages, proposed dedicated optimization with the source-side loss, used a different way to count model parameters, and designed different language tags for multilingual translation that could greatly affect zero-shot results~\citep{wu-etal-2021-language}. We leave the study of these differences to the future.

\paragraph{\lms variants do not offer better transfer characteristics for low-resource languages compared to \encdec.} One reason behind the popularity of multilingual NMT is its transfer capability to low-resource languages. We analyze this transfer behavior for \lms and explore transfer (to De) from similar (Fr) and distant (Zh) languages separately. Figure \ref{fig:transfer_high_to_low} shows the results. \encoderonly produces comparable results to \encdec, while \decoderonly lags far behind, and the incorporation of source-side objective actually hurts translation. Overall, we observe that \encdec almost dominates the transfer performance under different model sizes, regardless of language similarity. Similar results are also observed for low-resource to high-resource transfer (see Figure \ref{fig:transfer_low_to_high} in the Appendix).

\begin{table*}[t!]
\centering
\small
\caption{\label{tb:opus_100_results} Translation quality of different models for En$\rightarrow$XX,  XX$\rightarrow$En and zero-shot language pairs on OPUS-100. Models are trained in the Transformer big setting, aligned with 14-layer \encdec, containing about 412M parameters (excluding embedding and softmax layers). During training, we perform oversampling with a temperature of 5. We list average BLEU for High, Med, Low and All language groups. We also show average BLEU and translation language accuracy (ACC) for zero-shot test sets. 
}
\begin{tabular}{llrrrrrrrrrr}
\toprule
\multicolumn{2}{c}{\multirow{2}{*}{Model}}
    & \multicolumn{4}{c}{En$\rightarrow$XX} & \multicolumn{4}{c}{XX$\rightarrow$En} & \multicolumn{2}{c}{Zero-Shot} \\
  \cmidrule(lr){3-6} \cmidrule(lr){7-10} \cmidrule(lr){11-12}
  & & High & Med & Low & All & High & Med & Low & All & BLEU & ACC  \\
\midrule
\multicolumn{2}{c}{\encdec} & 25.8 & 32.4 & 31.9 & 29.2 & 31.4 & 34.3 & 35.0 & 33.1 &  4.80 & 24.21 \\
\midrule
\multirow{4}{*}{Deep}
& \encoderonly & -0.34 & -0.21 & -0.82 & -0.41 & -0.27 & -0.74 & -1.59 & -0.70 & 7.95 & 41.46 \\
& \quad + TopOnly & -0.01 & -0.14 & -1.79 & -0.44 & -0.07 & -0.71 & -1.43 & -0.57 & 6.59 & 39.06 \\
& \decoderonly & -4.51 & -8.18 & -12.9 & -7.47 & -5.18 & -10.1 & -13.0 & -8.38 & 4.10 & 25.60 \\
& \quad + TgtOnly & -0.83 & -0.78 & -1.40 & -0.93 & -1.27 & -1.81 & -2.43 & -1.69 & 7.34 & 39.62 \\
\midrule
\multirow{4}{*}{Wide}
& \encoderonly & -0.71 & -0.75 & -2.02 & -1.01 & -0.77 & -0.88 & -0.68 & -0.78 & 7.44 & 38.60 \\
& \quad + TopOnly & -0.40 & -0.37 & -0.66 & -0.45 & -0.47 & -0.50 & -1.41 & -0.69 & 6.92 & 37.69 \\
& \decoderonly & -4.25 & -7.58 & -12.2 & -7.03 & -5.05 & -9.88 & -13.3 & -8.32 & 4.49 & 28.08 \\
& \quad + TgtOnly & -1.29 & -1.27 & -0.82 & -1.18 & -1.88 & -1.96 & -2.04 & -1.94 & 5.53 & 29.75 \\
\bottomrule
\end{tabular}
\end{table*}

\paragraph{Comparison of \lms and \encdec variants on massively multilingual translation} We further examine the scalability of \lms with respect to the number of languages, and experiment on massively multilingual translation using OPUS-100. We enlarge the BPE size to 64K to handle multilingual lexicons. Following~\citet{zhang-etal-2020-improving}, we divide the test language pairs into high-resource (High, $>$0.9M), low-resource (Low, $<$0.1M), and medium-resource (Med, others) groups, and report average scores for each group. Table \ref{tb:opus_100_results} summarizes the results. \encdec outperforms \lms on supervised directions, with larger gap on low-resource languages and for XX$\rightarrow$En translation. By contrast, \lms, particularly \encoderonly, perform better on zero-shot directions, with improved translation language accuracy. Overall, {\encoderonly} outperforms \decoderonly, and also performs comparably to \encdec on supervised directions (often $<-1$ BLEU on average), echoing with our above findings.

\section{Conclusion and Discussion}

In this paper, we revisited language model architectures for machine translation from the perspective of model scaling and cross-lingual transfer. Extensive experiments show that \lms often have different scaling properties where the impact of architectural differences gradually reduce as models are scaled up, and that \lms often deliver better zero-shot transfer than its \encdec counterpart with improved off-target translation. While promising in zero-shot transfer, \lms lag behind the \encdec models in cross-lingual transfer for supervised directions. \encoderonly models with full visibility to the source input, show consistently outperform \decoderonly, and perform similarly well to \encdec across different settings. 
We expect that these findings could offer insights for researchers and practitioners focusing on developing new architectures, loss functions, regularizers or optimization methods for NMT.
Also, these findings show that while current product offerings for major language pairs or small on-device models should continue using \encdec, \lms can be an effective architecture for giant multilingual models with zero-shot transfer as a primary focus. 

The performance gap caused by architectural differences gradually disappear as the model sizes increase, with following implications: 1) Comparing NMT architectures only under one model setting (e.g. widely adopted 6-layer Transformer base) carries risks, because of the scaling properties difference. We recommend the best practice should portray the whole scaling picture for comparison. 2) Just like NMT models optimized for high-resource translation transfer poorly to low-resource scenarios, many models developed in the past with claims outperforming Transformer might not transfer to large-scale model settings and ideally should be revisited in the face of model scaling. 3) The off-target issue is one of the main bottlenecks for zero-shot translation, but why it happens and how to handle it without accessing (authentic or pseudo) training corpus on zero-shot directions still remains as an open questions. \encoderonly delivers promising zero-shot transfer, which deserves more attention. 


\clearpage

\bibliography{paper}
\bibliographystyle{icml2022}

\clearpage
\appendix

\section{Model Training and Inference}\label{app:train_inference}

We update model parameters via Adafactor~\citep{pmlr-v80-shazeer18a} with label smoothing of value 0.1, and scheduled learning rate of warmup steps 40K. We apply dropout of 0.1 to residuals, feed-forward activations and attentions. We employ the post-norm Transformer by default; for some exceptional cases (often with deep models where training is unstable) we use the pre-norm one instead. Batch size is set to about 128K tokens. We train models for up to 1M steps on different tasks, except Web En-De where 500K steps is used. We average 10 checkpoints for evaluation. For bilingual experiments, these checkpoints are selected according to the dev set performance; for multilingual experiments, we use the last 10 checkpoints. Beam search is used for inference, with a beam size of 8 and length penalty of 0.5. 

\section{More Experimental Results}

\begin{figure}[h]
    \centering
    \subcaptionbox{\label{fig:wmt_enfr_base_scaling_flops} En$\rightarrow$Fr}{
        \begin{minipage}[t]{0.38\textwidth}
        \centering
        \includegraphics[scale=0.38]{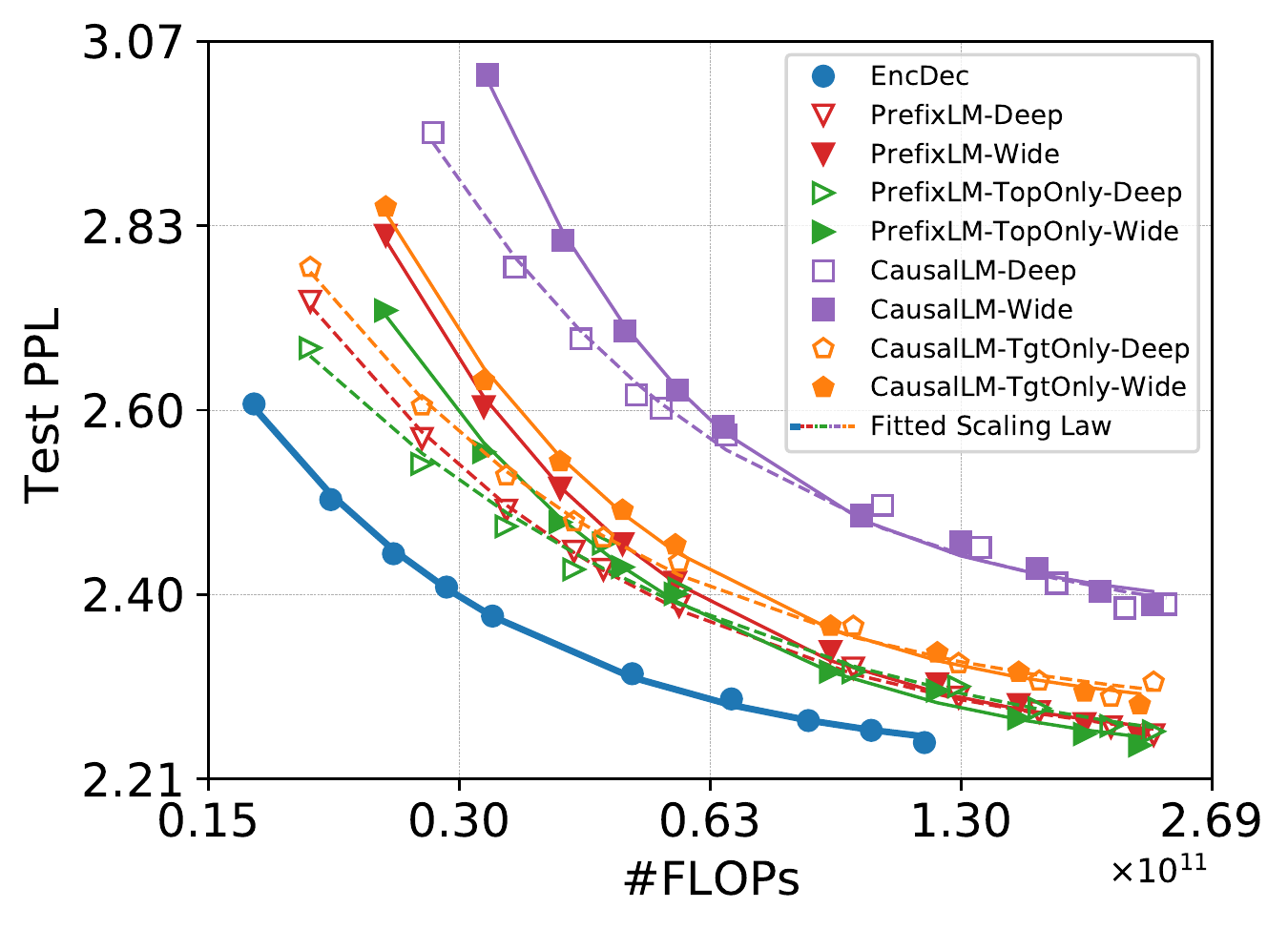}
        \end{minipage}%
    }
    \subcaptionbox{\label{fig:wmt_enzh_base_scaling_flops} En$\rightarrow$Zh}{
        \begin{minipage}[t]{0.38\textwidth}
        \centering
        \includegraphics[scale=0.38]{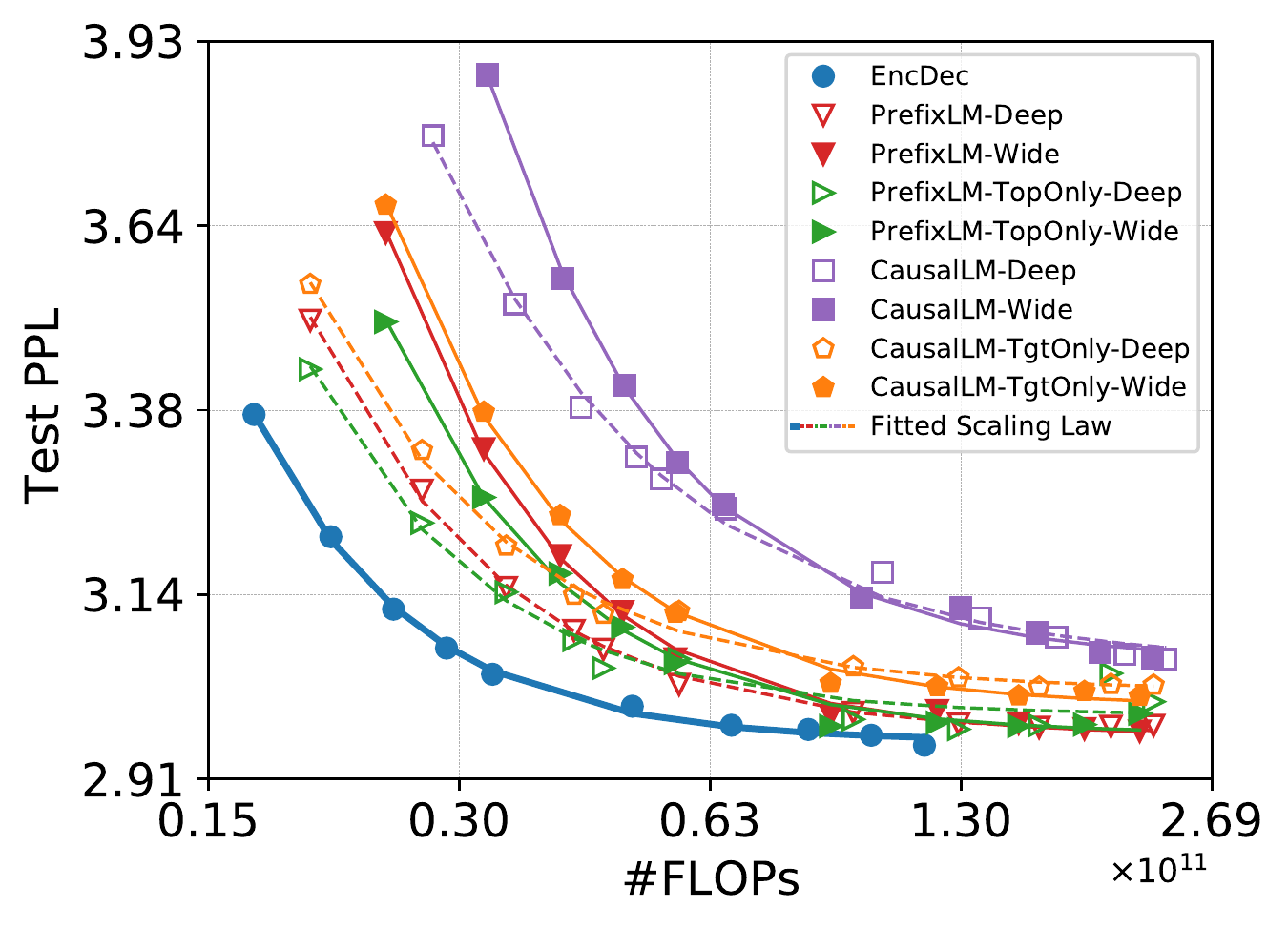}
        \end{minipage}%
    }
    \caption{\label{fig:base_scaling_flops} Fitted scaling curves for different models on WMT14 En-Fr and WMT19 En-Zh in term of FLOPs.}
\end{figure}

\begin{figure}[h]
    \centering
    \subcaptionbox{\label{fig:wmt_enfr_base_scaling_long} En$\rightarrow$Fr}{
        \begin{minipage}[t]{0.38\textwidth}
        \centering
        \includegraphics[scale=0.40]{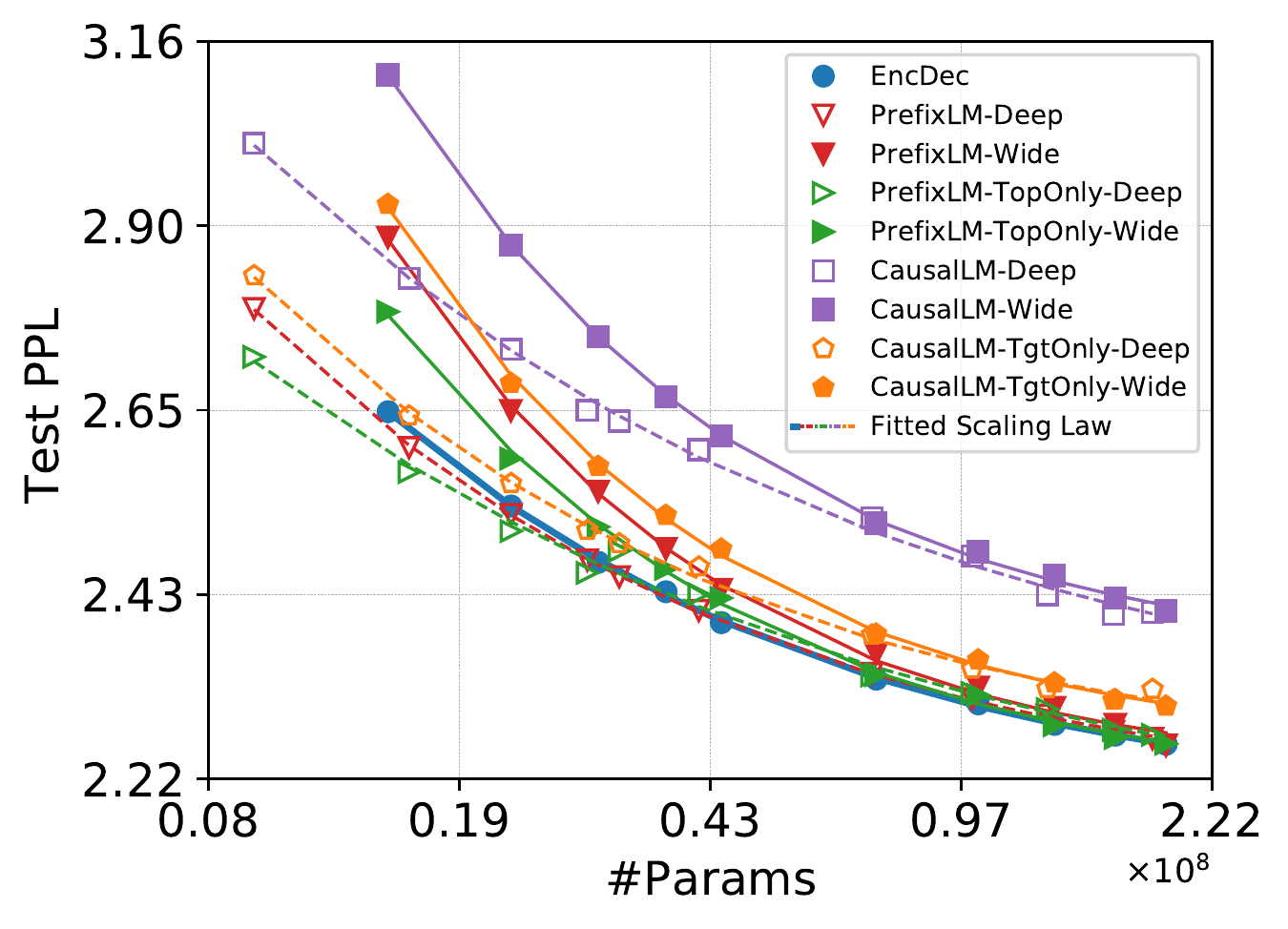}
        \end{minipage}%
    }
    \subcaptionbox{\label{fig:wmt_enzh_base_scaling_long} En$\rightarrow$Zh}{
        \begin{minipage}[t]{0.38\textwidth}
        \centering
        \includegraphics[scale=0.40]{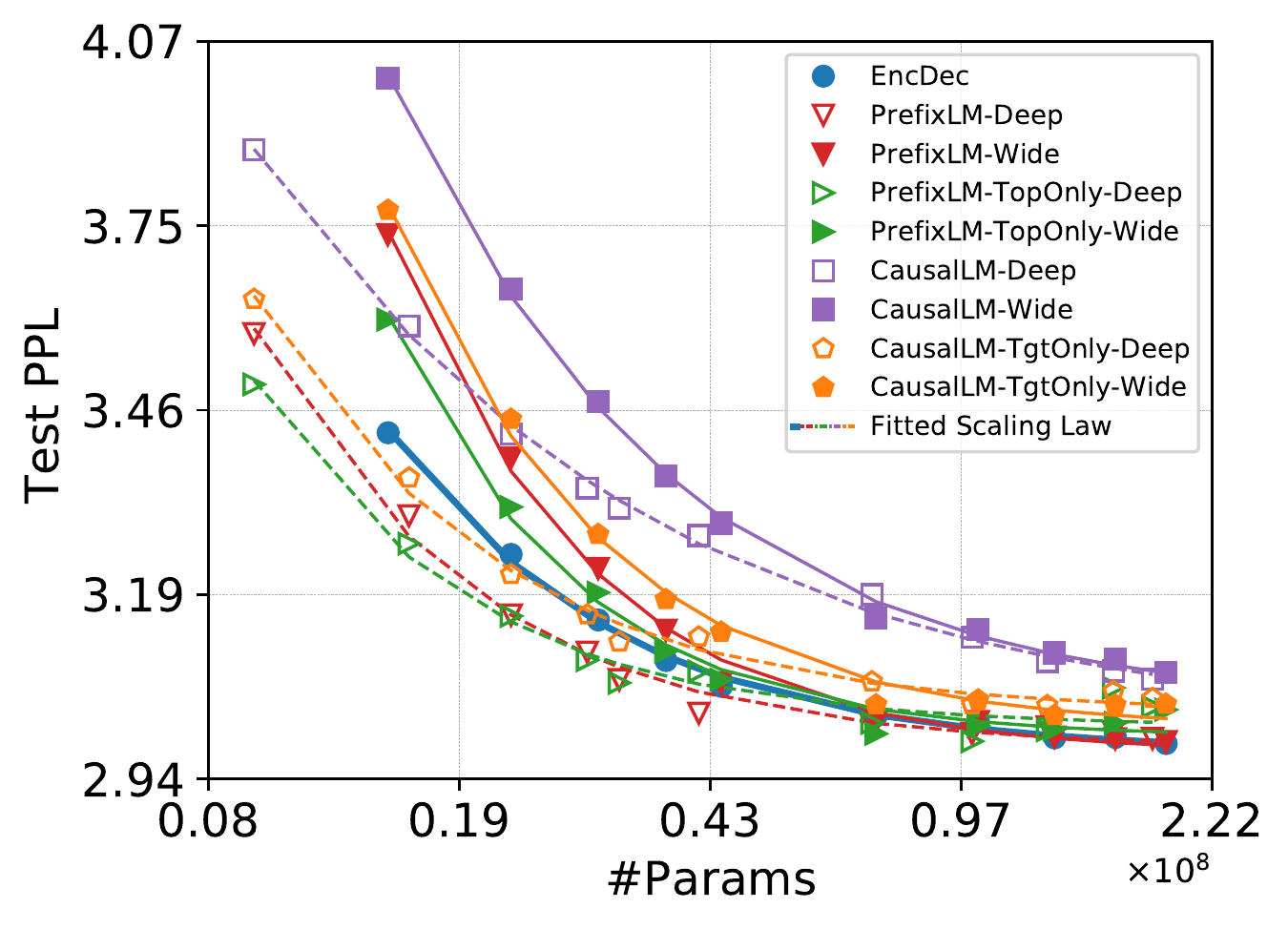}
        \end{minipage}%
    }
    \caption{\label{fig:base_scaling_long} Fitted scaling curves for different models on WMT14 En-Fr and WMT19 En-Zh on \textbf{the longest sentence group}. We rank our test set according to source sentence length, and then split it into 8 disjoint groups. This shows the results on the longest group.}
\end{figure}

\begin{figure}[h]
    \centering
    \subcaptionbox{\label{fig:wmt_enfr_base_scaling_layer} En$\rightarrow$Fr}{
        \begin{minipage}[t]{0.38\textwidth}
        \centering
        \includegraphics[scale=0.40]{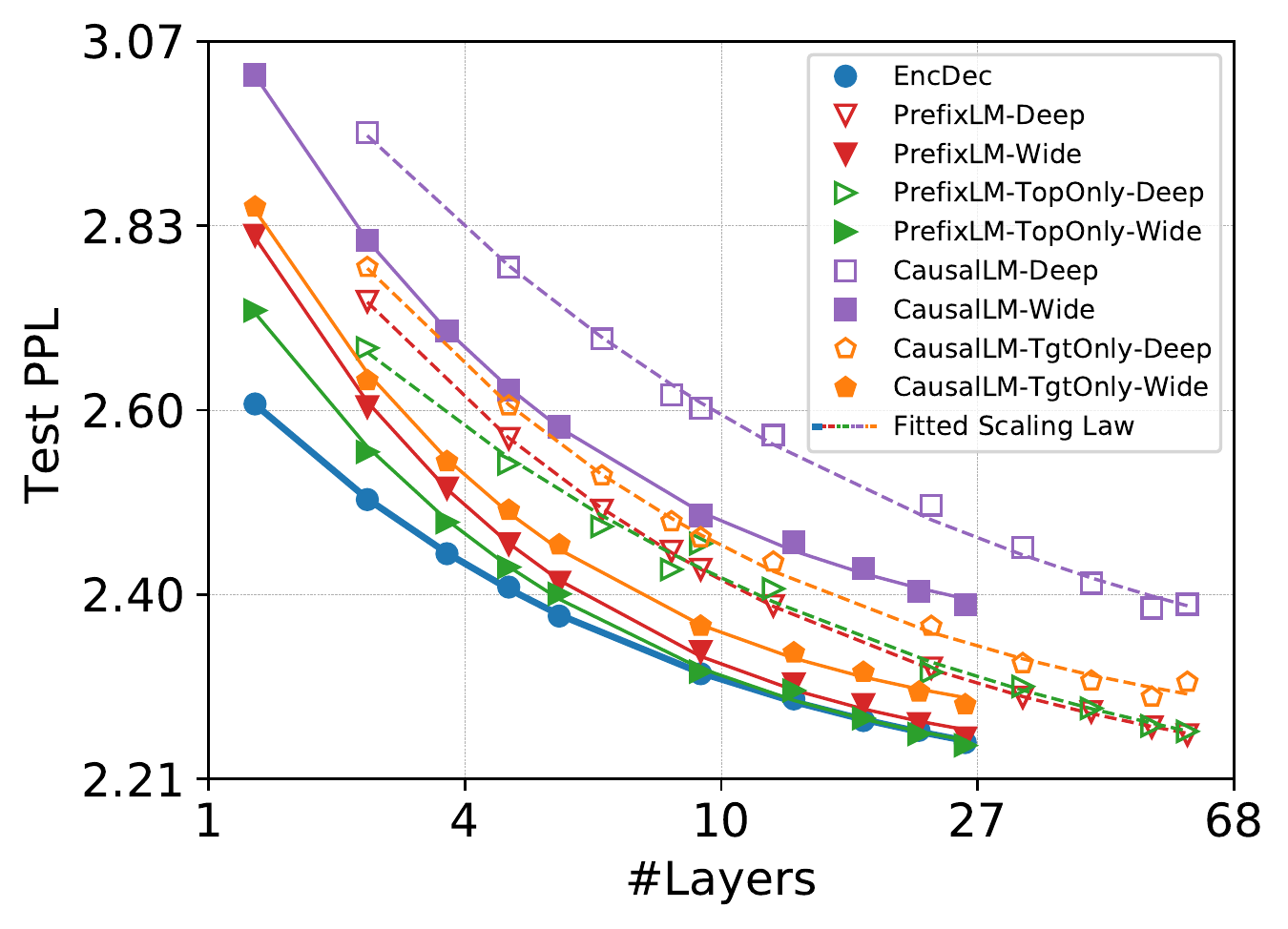}
        \end{minipage}%
    }
    \subcaptionbox{\label{fig:wmt_enzh_base_scaling_layer} En$\rightarrow$Zh}{
        \begin{minipage}[t]{0.38\textwidth}
        \centering
        \includegraphics[scale=0.40]{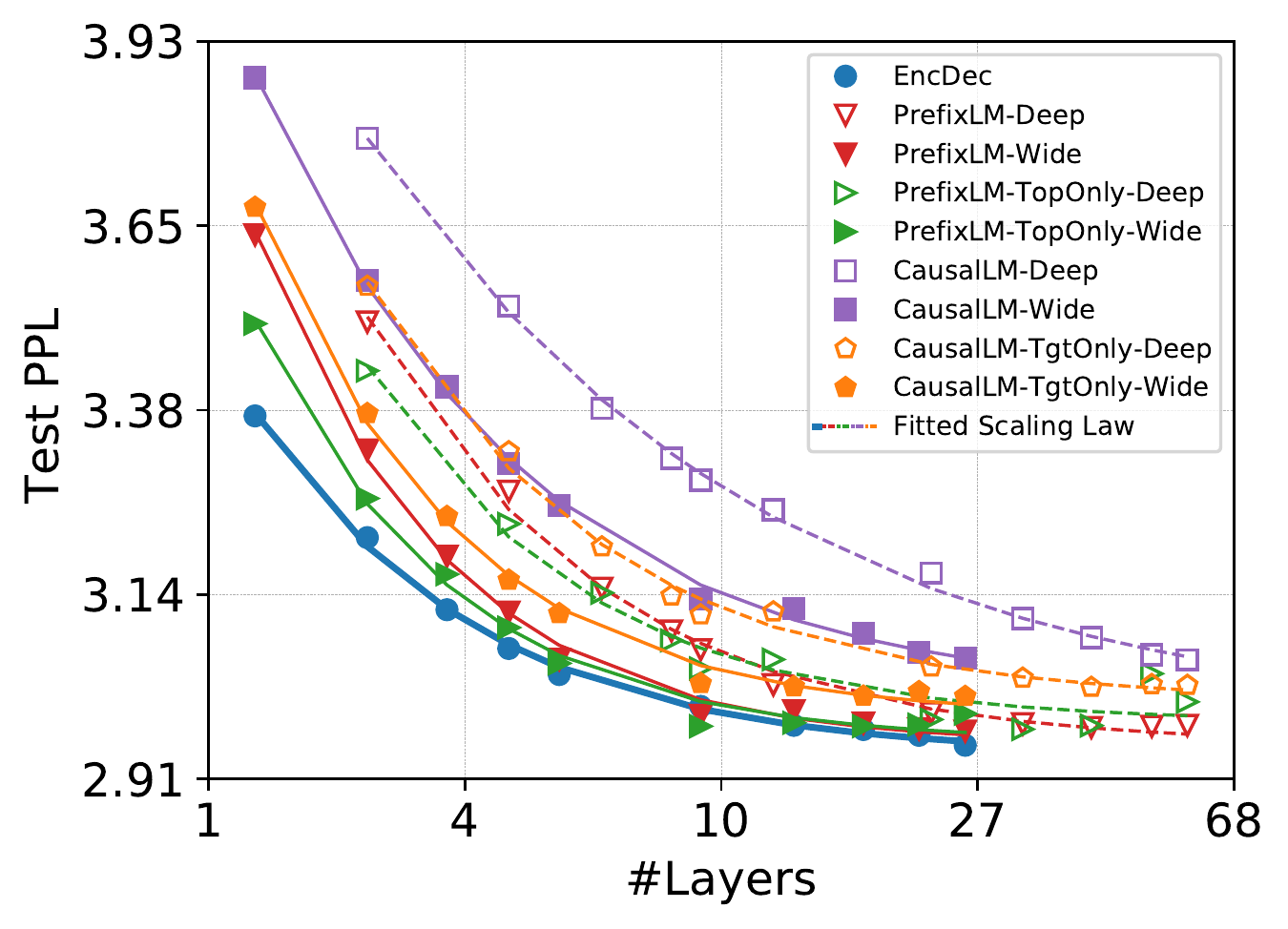}
        \end{minipage}%
    }
    \caption{\label{fig:base_scaling_layer} Fitted scaling curves for different models on WMT14 En-Fr and WMT19 En-Zh with respect to \textbf{the number of layers}. Note under the same number of layers, \textit{\lm + Deep} has much fewer parameters than \encdec and \textit{\lm + Wide}. The performance gap also narrows as model scales up.}
\end{figure}

\begin{figure*}[h!]
    \centering
    \subcaptionbox{\label{fig:wmt_enfr_base_scaling_srctgt} En$\rightarrow$Fr}{
        \begin{minipage}[t]{\textwidth}
        \centering
        \includegraphics[scale=0.40]{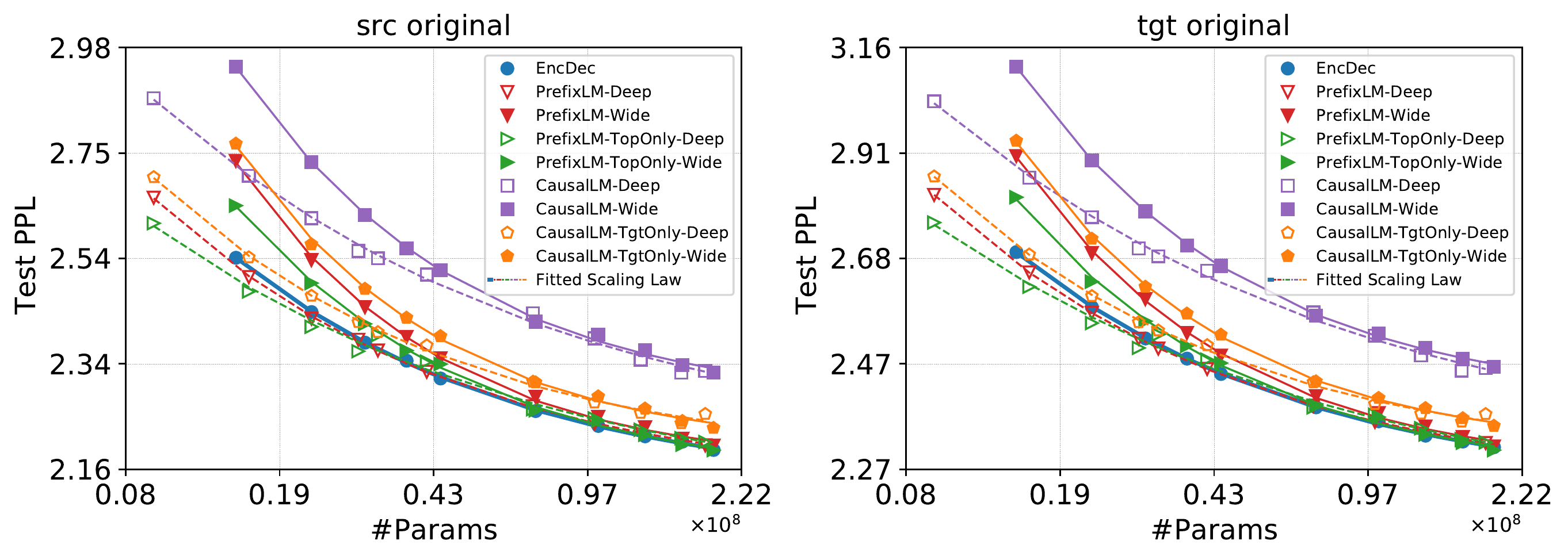}
        \end{minipage}%
    }\\
    \subcaptionbox{\label{fig:wmt_enzh_base_scaling_srctgt} En$\rightarrow$Zh}{
        \begin{minipage}[t]{\textwidth}
        \centering
        \includegraphics[scale=0.40]{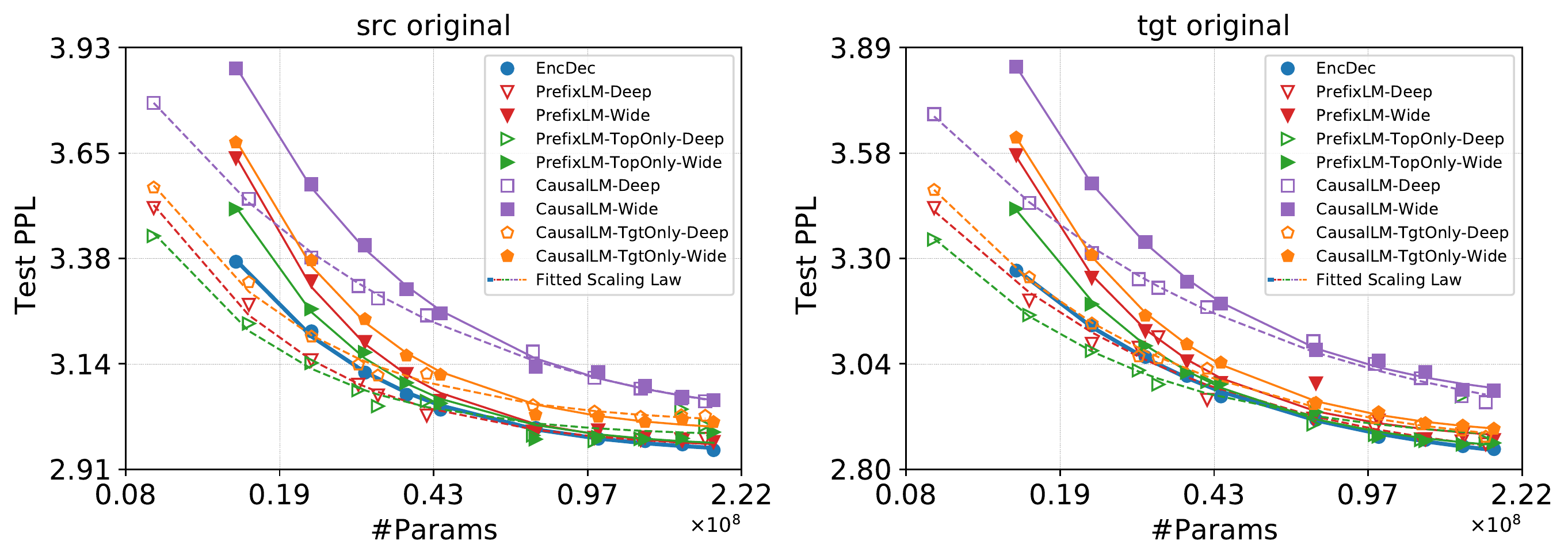}
        \end{minipage}%
    }
    \caption{\label{fig:base_scaling_srctgt} Fitted scaling curves for different models on WMT14 En-Fr and WMT19 En-Zh evaluated on \textit{source original} and \textit{target original} test sets.}
\end{figure*}

\begin{figure*}[t]
    \centering
    \subcaptionbox{\label{fig:wmt_enfr_base_scaling_614_bleu} En$\rightarrow$Fr}{
        \begin{minipage}[t]{\textwidth}
        \centering
        \includegraphics[scale=0.40]{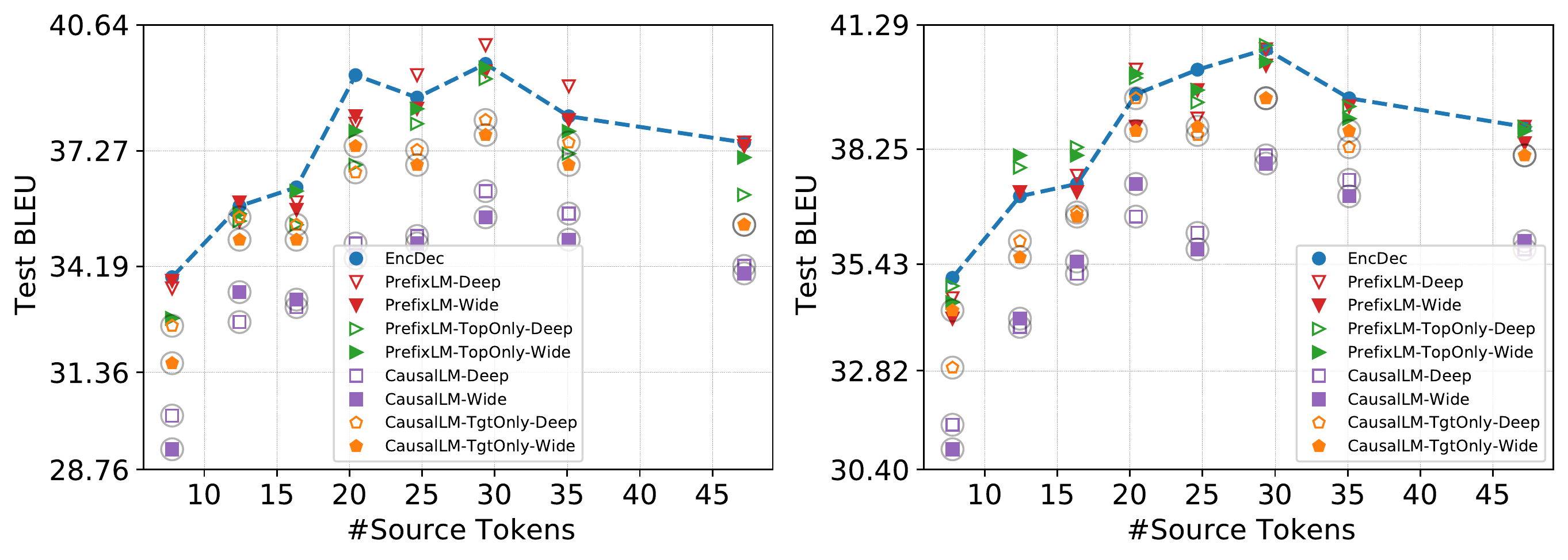}
        \end{minipage}%
    }\\
    \subcaptionbox{\label{fig:wmt_enzh_base_scaling_614_bleu} En$\rightarrow$Zh}{
        \begin{minipage}[t]{\textwidth}
        \centering
        \includegraphics[scale=0.40]{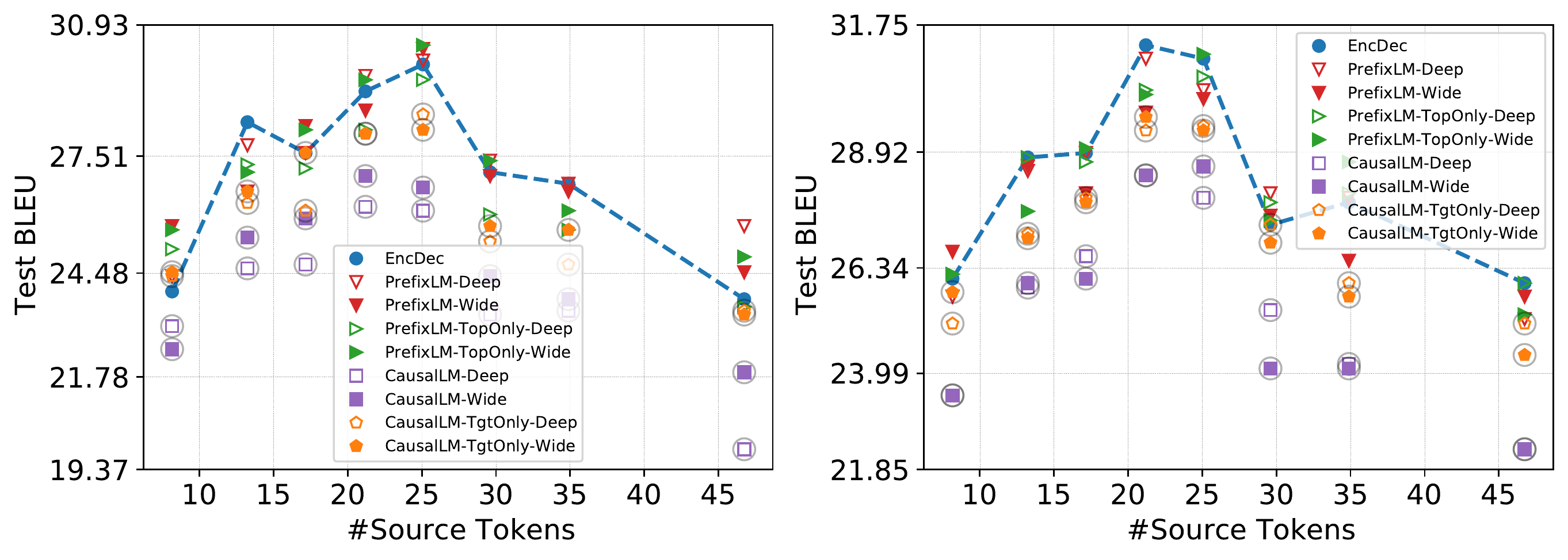}
        \end{minipage}%
    }
    \caption{\label{fig:base_scaling_614_bleu} BLEU scores for different models on WMT14 En-Fr and WMT19 En-Zh as a function of source sentence length. \textit{Left}: models aligned with 6-layer \encdec; \textit{Right}: models aligned with 14-layer \encdec.}
\end{figure*}

\begin{figure*}[t]
    \centering
    \includegraphics[scale=0.40]{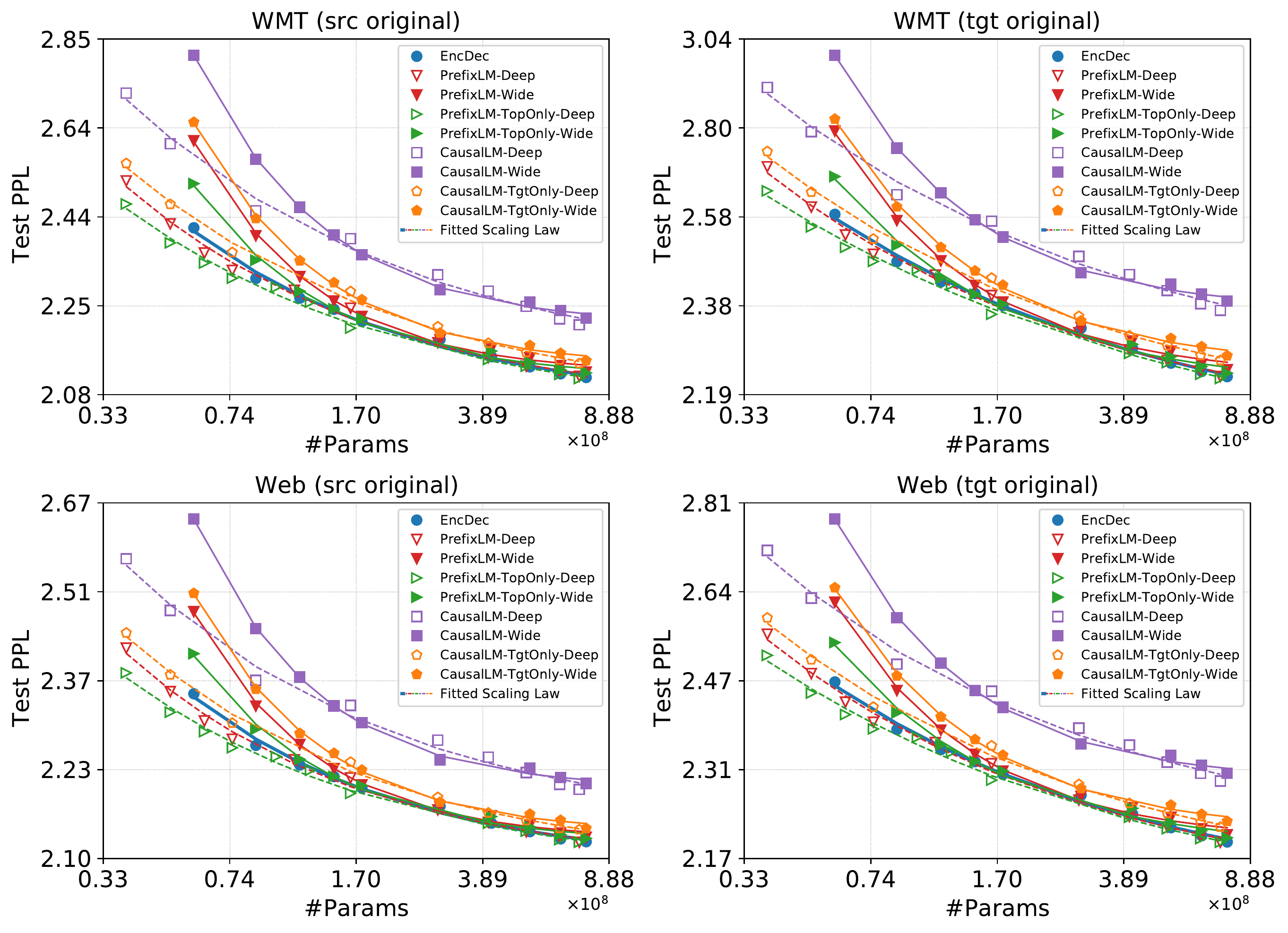}
    \caption{\label{fig:base_scaling_prod} Fitted scaling curves for different models on Web En-De (En$\rightarrow$De). \textit{src/tgt}: source/target; \textit{WMT}: out-of-domain evaluation set; \textit{Web}: in-domain evaluation set. Models are trained in the Transformer big setting.}
\end{figure*}

\begin{figure*}[t]
    \centering
    \includegraphics[scale=0.40]{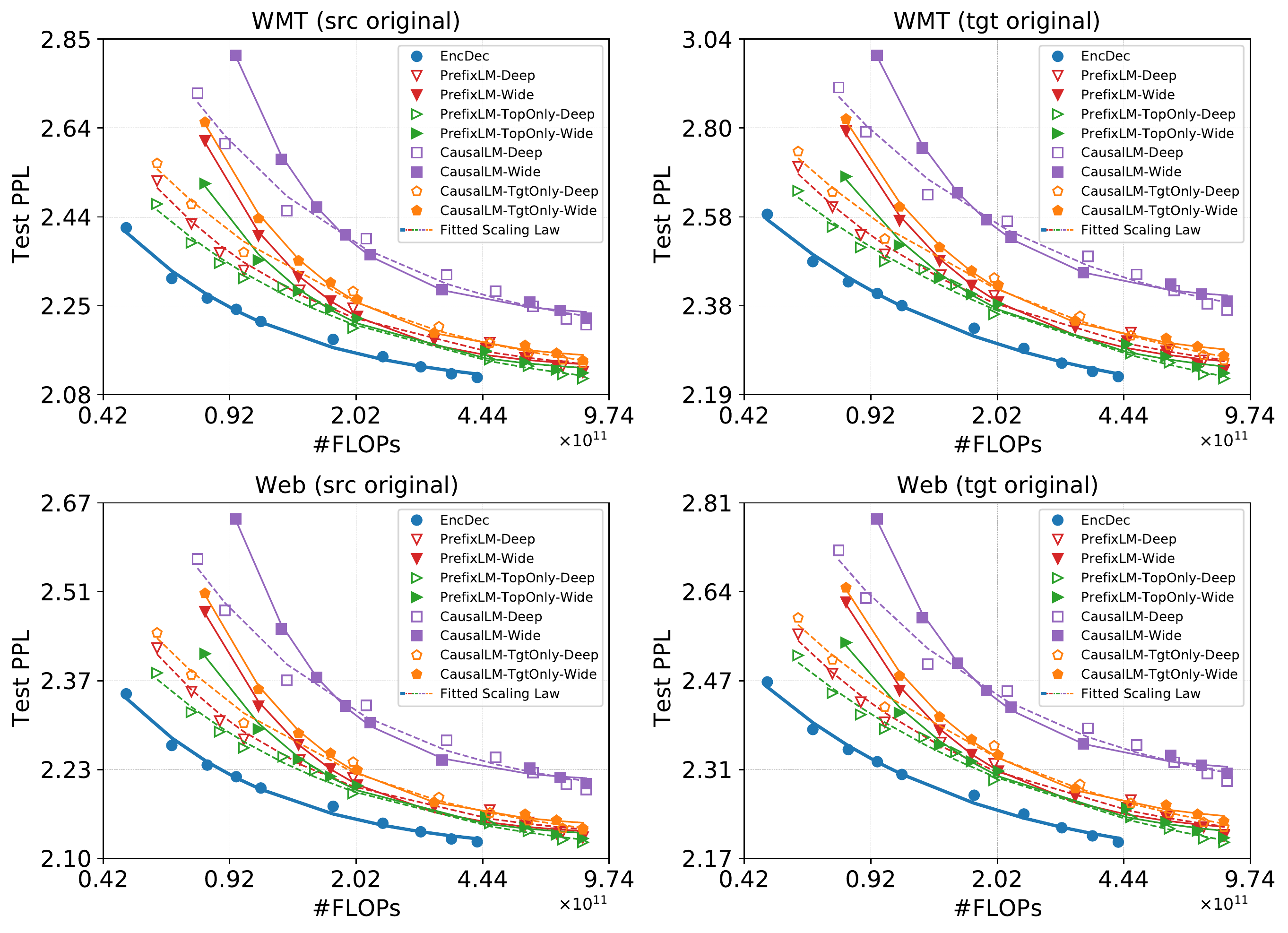}
    \caption{\label{fig:base_scaling_prod_flops} Fitted scaling curves for different models on Web En-De (En$\rightarrow$De) in terms of FLOPs. Models are trained in the Transformer big setting.}
\end{figure*}

\begin{figure*}[t]
    \centering
    \includegraphics[scale=0.40]{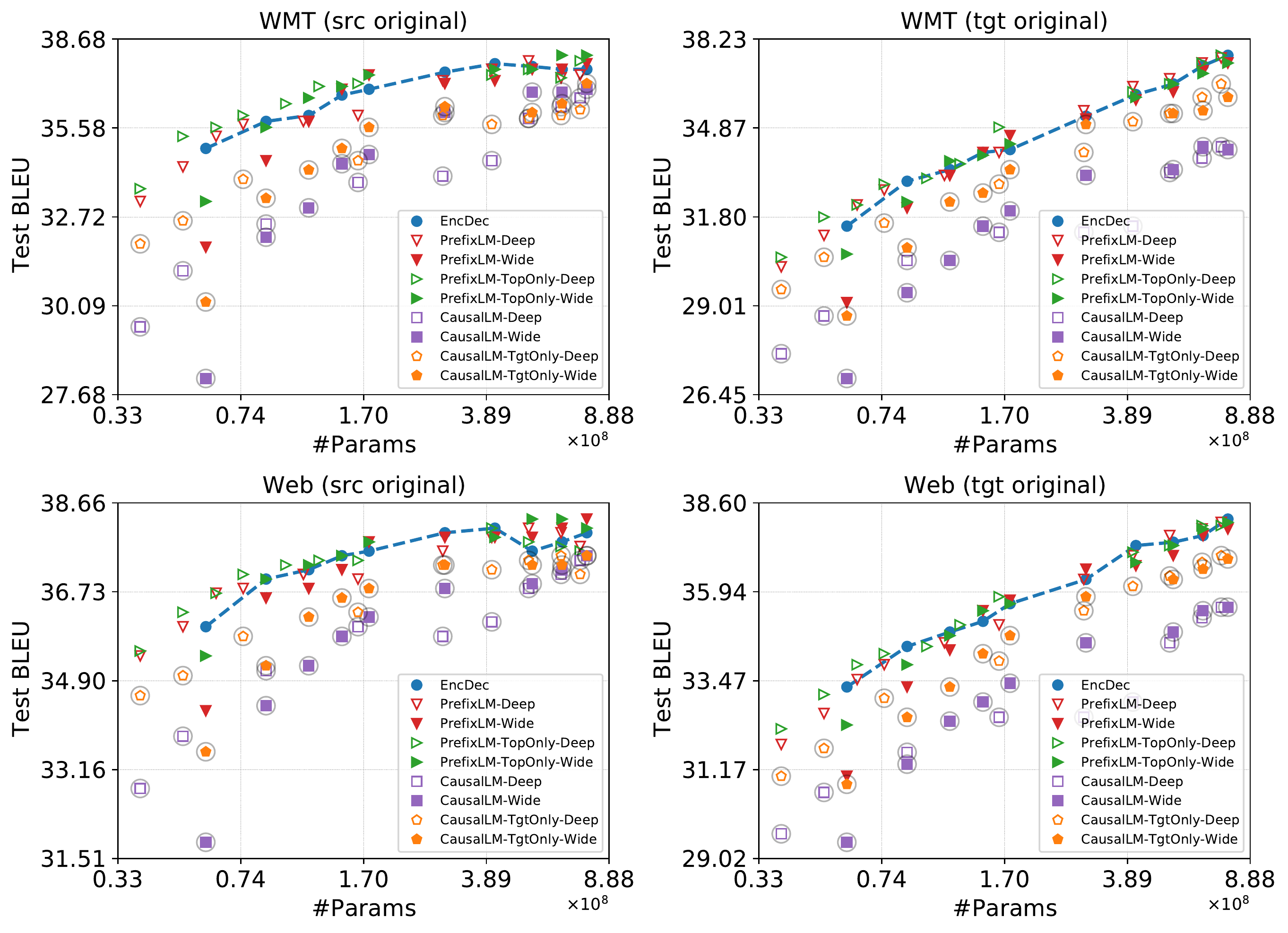}
    \caption{\label{fig:base_scaling_prod_bleu} BLEU scores for different models on Web En-De (En$\rightarrow$De) as a function of model parameters. Models are trained in the Transformer big setting.}
\end{figure*}

\begin{figure*}[t]
    \centering
    \includegraphics[scale=0.40]{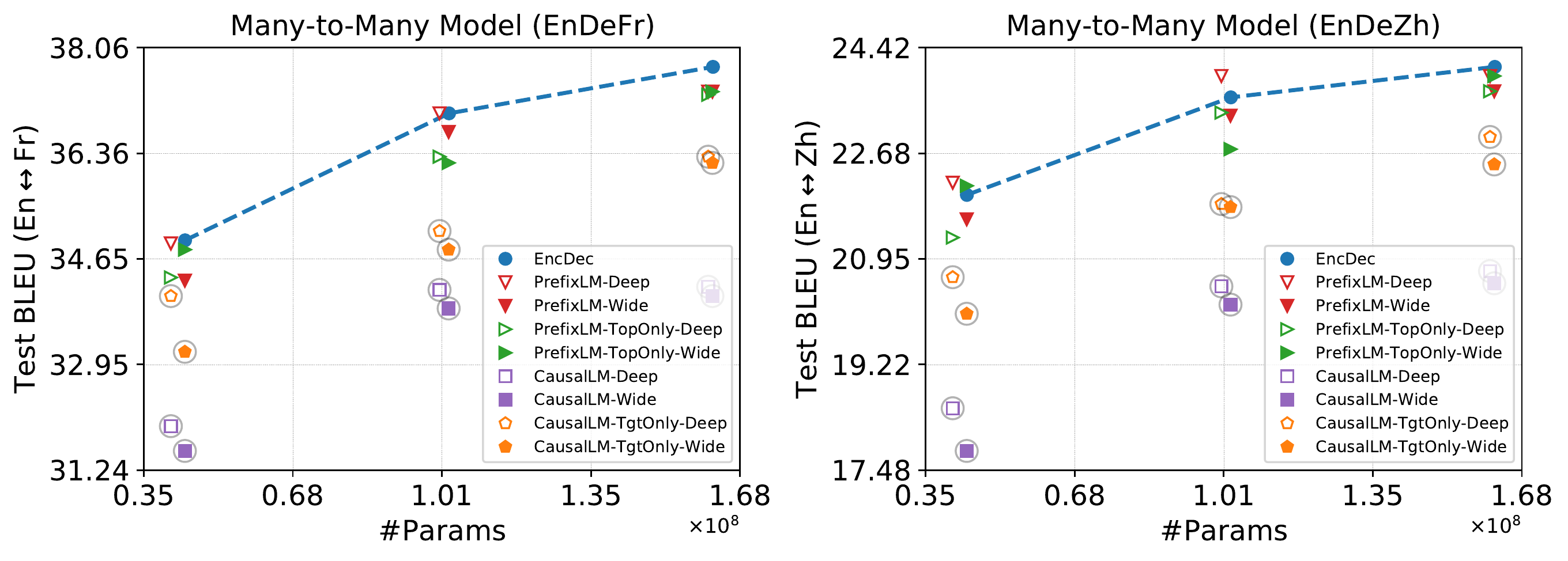}
    \caption{\label{fig:transfer_low_to_high} Cross-lingual transfer results (average BLEU scores) for different models from the low-resource language (En-De) to high-resource directions under different model sizes on WMT datasets. Average is performed over En$\leftrightarrow$Fr/Zh. \textbf{Left}: multilingual En-De-Fr system; \textbf{Right}: multilingual En-De-Zh system.}
\end{figure*}


\begin{figure*}[t]
    \centering
    \includegraphics[scale=0.40]{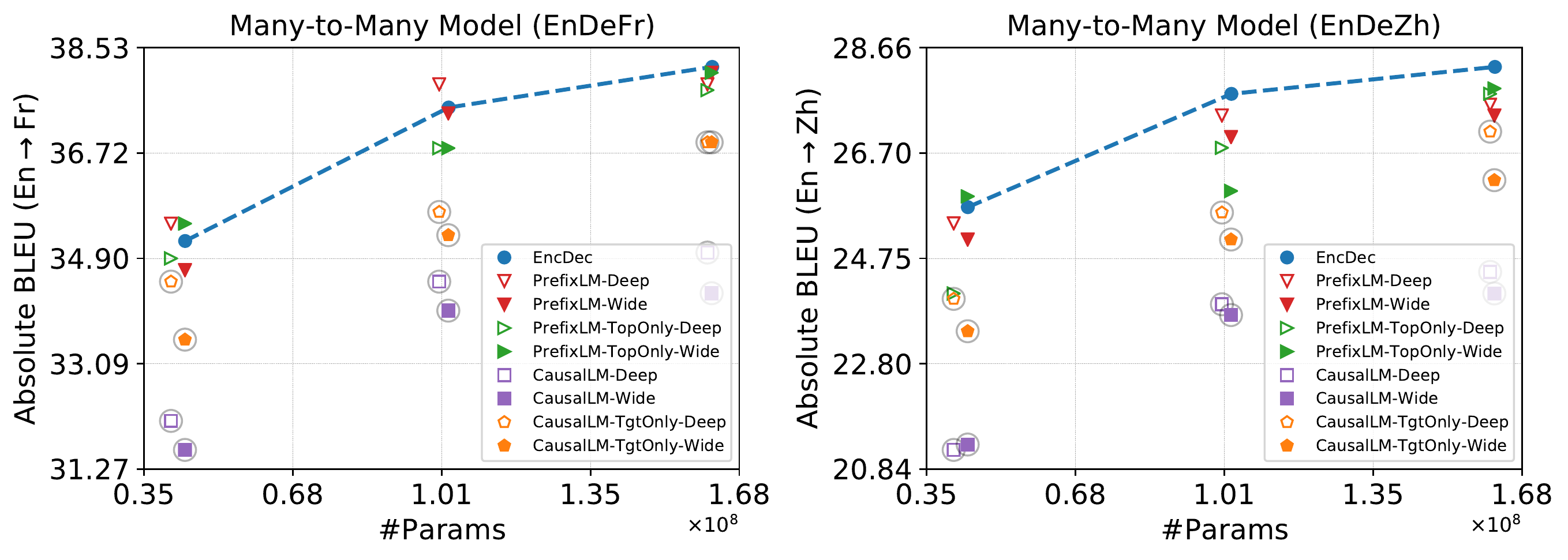}
    \includegraphics[scale=0.40]{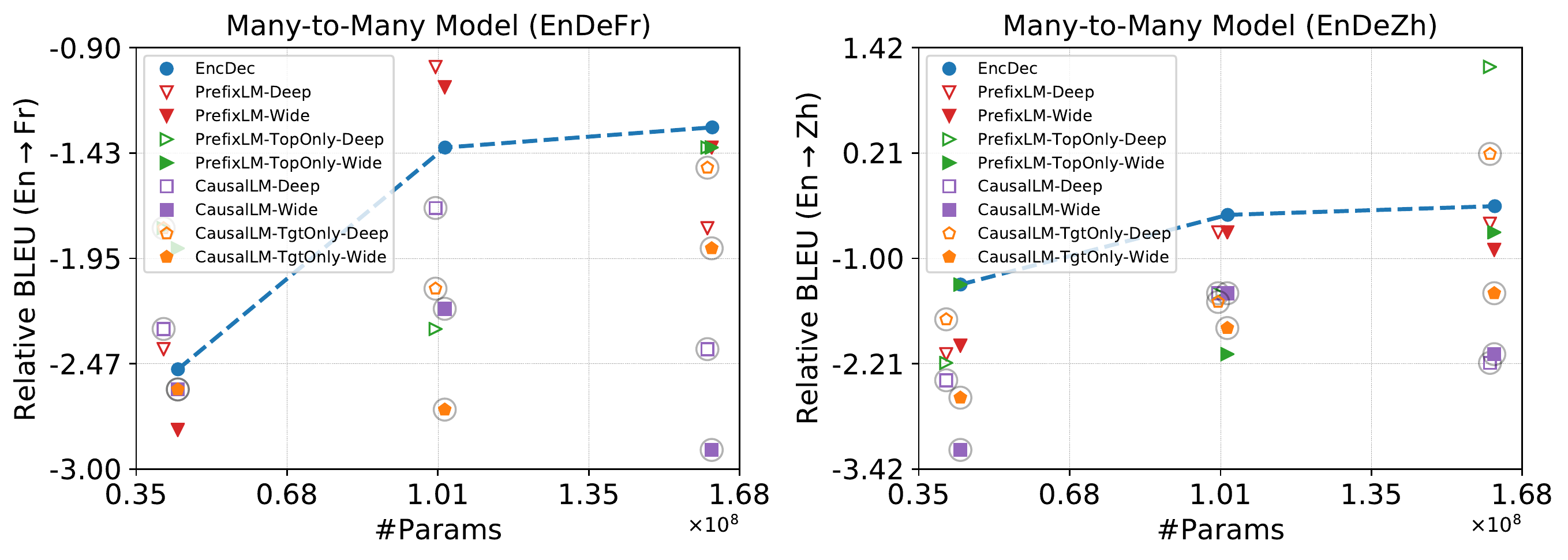}
    \caption{\label{fig:transfer_low_to_high_relabs} \response{Absolute (top) and relative (bottom) transfer results of different models for En$\rightarrow$Fr and En$\rightarrow$Zh under different models sizes on WMT datasets. \textbf{Left}: multilingual En-De-Fr system; \textbf{Right}: multilingual En-De-Zh system. Relative score is computed by comparing multilingual model and its corresponding bilingual counterpart. Overall, there is no clear pattern supporting that \lms encourage knowledge transfer better than \encdec.}}
\end{figure*}

\end{document}